\documentclass[preprint]{elsarticle}
\usepackage{amssymb}
\usepackage{amsmath}
\usepackage{booktabs}%
\usepackage{algorithm}
\usepackage{wrapfig}
\usepackage[noend]{algpseudocode}
\usepackage{graphicx}      
\usepackage{subcaption}  
\usepackage{microtype}
\usepackage{hyperref}
\begin{document}
\begin{frontmatter}

\title{Frequency-Aware Vision Transformers for High-Fidelity Super-Resolution of Earth System Models}

\author{Ehsan Zeraatkar*} 
\ead{ehsanzeraatkar@txstate.edu}

\affiliation{organization={Computer Science, Texas State University},
            city={San Marcos},
            state={Texas},
            country={US}}

\author{Salah A Faroughi} 

\affiliation{organization={Chemical Engineering, The University of Utah},
            city={Salt Lake City},
            state={UT},
            country={US}}

\author{Jelena Te\v{s}i\'c}

\affiliation{organization={Computer Science, Texas State University},
            city={San Marcos},
            state={Texas},
            country={US}}

\begin{abstract}
Super-resolution can play an essential role in enhancing the spatial fidelity of Earth System Model outputs, allowing fine-scale structures highly beneficial to climate science to be recovered from coarse simulations. However, traditional deep \emph{super-resolution} methods, including convolutional and transformer based models, tend to exhibit spectral bias, reconstructing low-frequency content more readily than valuable high-frequency details. In this work, we introduce ViSIR and ViFOR, two frequency-aware frameworks. ViSIR stands for the Vision Transformer-Tuned Sinusoidal Implicit Representation. ViSIR combines vision transformers with sinusoidal activations to mitigate spectral bias. ViFOR stands for the Vision Transformer Fourier Representation Network. ViFOR integrates explicit Fourier based filtering for independent low- and high-frequency learning. Evaluated on the E3SM-HR Earth system dataset across surface temperature, shortwave, and longwave fluxes, these models outperform leading Convolutional NN, Generative Networks, and vanilla transformer baselines, with ViFOR demonstrating up to 2.6~dB improvements in Peak Signal to Noise Ratio and higher Structural Similarity. 

\end{abstract}

\begin{keyword}
Earth System Models, Super-Resolution, Vision Transformers, Implicit Neural Representations, Spectral Bias, Climate Data
\end{keyword}
\end{frontmatter}

\section{Introduction}
\label{sec-intro}

The accelerating impacts of climate change demand ever more precise, fine-grained predictions of atmospheric and Earth-system behavior to support monitoring, hazard assessment, and climate-adaptation decisions. Earth System Models (ESMs) are the principal scientific tools for simulating coupled interactions among the atmosphere, oceans, land, ice, and biosphere, and they underpin our understanding of global and regional climate variability~\cite{collins2006}. However, the computational cost of running these models at truly high spatial resolution remains prohibitive, so even state-of-the-art ESMs typically produce outputs on coarse grids that lack the local detail needed for many regional geosensing applications and decision-making contexts~\cite{heinze2019}. As emphasized in the CAIG program, overcoming this resolution gap is critical for expanding the practical use of geoscience data and models in societally relevant settings, from water-resource management to hazard early warning.  

This critical gap motivates the development of super-resolution methods that transform coarse ESM outputs into high-fidelity, high-resolution fields~\cite{eyring2016}. The \emph{super-resolution} problem—reconstructing high-resolution (HR) images from low-resolution (LR) inputs—has become a core challenge in computer vision~\cite{rahaman2019}, with broad impact on medical imaging~\cite{Bashir2021}, wildlife tracking~\cite{kuzmanic2007}, security and surveillance~\cite{Grosche2021}, cultural preservation, and consumer electronics~\cite{Maral2022}. In all of these domains, \emph{super-resolution} methods must navigate the fundamental uncertainty of inferring missing fine-scale information from coarse inputs while balancing computational tractability against accurate restoration of small-scale structure~\cite{ledig2017, lim2017}. For ESM data, this challenge is amplified: fine gradients, localized extremes, and sharp fronts in climate variables render naive downscaling or generic image models inadequate for regional climate applications and geosensing tasks~\cite{vandal2017}, motivating specialized \emph{super-resolution} algorithms that are both frequency-aware and physically meaningful.  

Crucially, learning based super-resolution does not directly recover unresolved physical processes or dynamically generate fine-scale information from a single low-resolution snapshot. Instead, it follows a statistical learning paradigm in which paired low- and high-resolution data are used during training to learn conditional relationships between large-scale patterns and associated small-scale variability. The reconstructed high-resolution fields therefore encode statistically learned fine-scale structures conditioned on coarse-scale inputs, rather than explicitly simulated dynamics. In the context of Earth system models, much of the apparent fine-scale variability reflects persistent, unresolved influences such as topography, land–surface heterogeneity, or boundary forcing, whose signatures are implicitly encoded through the training data even in the absence of explicit high-resolution auxiliary predictors. Consequently, the proposed method should be understood as a statistical downscaling technique rather than a form of dynamical downscaling, providing an efficient pathway to enhance the usability of ESM outputs for geosensing and decision support.  

While deep learning methods—especially convolutional neural networks (CNNs) and, more recently, Vision Transformers (ViTs)—have produced impressive advances in \emph{super-resolution} for natural imagery, they face notable limitations when applied to Earth system data. A persistent challenge is \textit{spectral bias}, whereby deep networks preferentially capture smooth, low-frequency structures and struggle to reconstruct the high-frequency components that encode physically meaningful gradients, fronts, and localized extremes in climate fields. This mismatch between model bias and the multiscale, structured variability of ESM outputs hinders the recovery of geophysically relevant small-scale information essential to many geosensing applications.  

Recent breakthroughs in deep learning have nonetheless yielded powerful \emph{super-resolution} models, including CNNs~\cite{Kim2016VDSR}, ViTs~\cite{Dosovitskiy2020}, and generative diffusion techniques~\cite{Dickstein2015}. Vision Transformers are particularly promising because they can model long-range spatial dependencies, a key requirement for geophysical fields with large-scale connections and coherent structures. Nevertheless, even Transformer based methods continue to struggle with recovering sharp high frequency details due to \textit{spectral bias} \cite{bai2022improving}. To address this limitation, we first developed the \emph{Vision Transformer Tuned Sinusoidal Implicit Representation Network (ViSIR)}, which combines the global contextual modeling capabilities of Vision Transformers with sinusoidal activations in an implicit neural representation (INR) decoder. Guided by ViT-extracted features, ViSIR enhances reconstruction of high-frequency details and partially mitigates spectral bias; however, consistently balancing low- and high-frequency content across heterogeneous climate variables remains challenging.

To more directly confront this problem, we introduce a frequency-aware framework for climate super-resolution that leverages the global modeling strengths of ViTs while explicitly suppressing spectral bias. Our method unfolds in two stages: first, integrating sinusoidal activations into transformer architectures in ViSIR; and second, extending to explicit Fourier-based frequency separation in the Vision Transformer Fourier Representation Network (ViFOR). These architectures substantially improve reconstruction of high-frequency, physically meaningful gradients across key Earth system variables, thereby enhancing the spatial fidelity and practical value of ESM outputs for geosensing and climate services.  

\textbf{Domain-Specific Motivation and Novelty.} ESM variables exhibit characteristic spatial structure: large-scale, smoothly varying background fields coexist with localized sharp gradients, fronts, and discontinuities associated with topography, cloud systems, and surface fluxes. While recent INR+Transformer architectures have been explored in natural image processing and generic scientific computing, they have not been designed to account for these multiscale geophysical characteristics or the operational needs of climate and geosensing applications. ViSIR and ViFOR explicitly tailor frequency-aware implicit representations to the spatial statistics of ESM data, enabling scale-aware reconstruction that preserves both global coherence and localized physical variability. To the best of our knowledge, this work represents the first systematic, frequency-explicit integration of implicit neural representations and Vision Transformers, tailored explicitly to Earth system model super-resolution, with an architectural design and evaluation grounded in the multiscale behavior of climate variables and the broader goal of expanding the actionable use of geoscience data.  

\noindent\textbf{Contributions: } This paper introduces \textbf{ViSIR}, a hybrid Vision Transformer and updated SIREN architecture that integrates sinusoidal activations directly within the Transformer encoder to mitigate spectral bias during global contextual feature learning for ESM super-resolution. We further present \textbf{ViFOR}, which extends ViSIR by incorporating Fourier-based activations embedded inside Transformer blocks, enabling explicit decomposition of low- and high-frequency components rather than relying on a single global frequency parameter. The evaluation presented on the E3SM-HR dataset demonstrates consistent and robust improvements in Peak Signal-to-Noise Ratio, Structural Similarity, and Mean Squared Error relative to CNN-, ViT-, Transformer-, and GAN-like baselines. Comprehensive ablation studies quantify the effect of sinusoidal frequencies in ViSIR and Fourier cutoffs in ViFOR, clarifying their strengths and limitations. Notably, ViFOR benefits markedly from full-image training, revealing the critical role of global context and establishing a scalable framework for ESM downscaling. ViSIR represents an initial step toward frequency-aware transformer-based super-resolution by mitigating spectral bias within attention -based feature learning through sinusoidal activations. Building on this foundation, ViFOR introduces explicit Fourier-based frequency separation to independently model low- and high-frequency components, providing a more targeted and stable solution for ESM variables with heterogeneous spectral characteristics and offering a practical, frequency-aware pathway to enhance the geosensing utility of coarse-resolution climate model outputs.

\section{Related Work}
\label{sec-related}

Super-resolution has been a computer vision challenge over the years, extending from remote sensing and medical imaging to environmental and climate sciences. \emph{super-resolution} is either single-image SR (SISR), in which one low-resolution (LR) input is restored into one high-resolution (HR) image, or multi-image SR (MISR), in which complementary information across a sequence of LR observations is used to estimate a better HR reconstruction \cite{Maral2022}. While classical methods, such as reconstruction- and example based schemes \cite{Yang2010}, yielded early improvements, they were limited by their reliance on strong assumptions about image structure and smoothness, which made them unsuitable for general real-world cases.

Convolutional Deep Neural Networks (CNNs) have revolutionized the super -resolution task, and the seminal super-resolution convolutional neural network, or SRCNN \cite{Dong2016}, brought end-to-end mapping from LR to HR images, beating interpolation-based baselines. Subsequent models, such as VDSR \cite{Kim2016VDSR}, EDSR \cite{Kim2016EDSR}, and RDN \cite{Zhang2018}, delved deeper into network architectures and employed residual and dense connections to improve the reconstruction of fine-grained details. However, CNN-based architectures exhibit a pronounced tendency to learn low-frequency, smoothly varying features earlier and more easily than high-frequency components, a phenomenon commonly referred to as \emph{spectral bias} \cite{Zhang2019}. This bias is particularly detrimental to Earth System Model (ESM) applications where high-contrast temperature and flux gradients in spatially localized areas harbor critical scientific information.

To address these problems, implicit neural representations (INRs), directly mapping continuous spatial coordinates to signal values, have been explored. Recent INR-based super-resolution methods have also been studied in remote sensing image fusion, including a Mamba-based collaborative implicit neural representation framework for hyperspectral \cite{Mamba2025} and multispectral fusion and a spatial–frequency dual-domain implicit guidance method based on Kolmogorov –Arnold networks \cite{ZHU2025}, which jointly exploit spatial continuity and frequency-domain priors to enhance high-resolution reconstruction. The Sinusoidal Representation Network (SIREN) \cite{SIREN} demonstrated periodic activation functions that can handle spectral bias effectively, enabling neural networks to more accurately represent high-frequency components that are present in the training data but are often underfit by standard activation functions. Subsequent works such as Generalized INRs (GINR) \cite{grattarola2022ginr} and Higher-Order INRs (HOIN) \cite{chen2024hoin} extend these ideas by incorporating spectral graph embeddings and neural tangent kernels. In parallel, generative adversarial techniques, such as SRGAN \cite{shidqi2023}, aim to achieve perceptual realism and have been applied to downscale coarse-resolution climate data.
In contrast, multimodal architectures that combine U-Nets with attention modules have been used to improve regional temperature forecasts \cite{Ding2024}. NeurOp-Diff \cite{neurOpDiff2025} introduced a neural operator diffusion framework for continuous-scale super-resolution, demonstrating strong generalization to arbitrary scale factors with publicly available code. Similarly, PC-SRGAN \cite{pcSRGAN2025} proposed a physically consistent SRGAN for transient scientific simulations, enforcing conservation properties while enhancing reconstruction quality. Despite their successes, these methods either rely on localized convolutional inductive biases, focus on perceptual realism or physical constraints, or lack explicit mechanisms for global frequency separation and scalable representation learning across heterogeneous climate variables.

Vision Transformers (ViTs) \cite{Dosovitskiy2020} have recently appeared as serious alternatives to the conventional CNNs in \emph{super-resolution} tasks. By partitioning images into patches and applying multi-head self-attention, ViTs excel at modeling long-range spatial relations and global context. This makes them particularly attractive for ESM datasets, where global-scale behavior must be preserved. However, ViTs also exhibit spectral bias, which means they collapse fine details and high-frequency features across all tasks \cite{bai2022improving}. Efforts to simplify transformer-based \emph{super-resolution} models \cite{Zhong2023, Karwowska2024} have improved efficiency but have not resolved the inherent problem of frequency imbalance. Most recently, TTRD3 \cite{ttrd32025} combined texture transfer and dual diffusion modeling to recover fine-grained remote-sensing textures, underscoring the trend toward \emph{super-resolution} methods that incorporate frequency and texture information. 

These developments directly motivate our contributions. First, we introduced the \emph{Vision Transformer–Tuned Sinusoidal Implicit Representation Network (ViSIR)}, which integrates ViTs and sinusoidal INRs to mitigate spectral bias and recover fine details in ESM data. While effective, ViSIR still falls short of balancing high- and low-frequency learning for heterogeneous climate variables. From this, we propose the \emph{Vision Transformer Fourier Representation Network (ViFOR)}, which expands the method by adding Fourier based activation functions. The design introduces an explicit separation of low- and high-frequency components within a Transformer-guided implicit representation framework, providing a principled mechanism to address spectral imbalance in ESM super-resolution.

The downscaling of global climate model outputs has a relatively long and well-established history within the climate science community and is generally classified into two categories: dynamical and statistical methods \cite{hewitson1996climate}. Dynamical downscaling is based on nesting high-resolution regional climate models within coarse-resolution global models. These provide physically consistent simulations at increased computational cost. On the contrary, classical statistical downscaling methods include regression based techniques, weather typing, bias correction, and analog methods, and geostatistical simulation—learn empirical relationships between large-scale predictors and local-scale variables and have been widely applied for climate impact assessment and water resource studies \cite{hewitson2014interrogating,lantuejoul2002geostatistical}. All learn empirical relationships between large-scale predictors and local-scale variables. These offer computational efficiency but often have limited ability to recover fine-scale spatial structure. The methods proposed in this work, namely ViSIR and ViFOR, are complementary to these established paradigms, positioning data-driven super-resolution as a frequency-aware extension of statistical downscaling rather than a replacement for physical or dynamical models. None of these replace physical or classical statistical downscaling; instead, they leverage modern representation learning that explicitly addresses spectral bias and multiscale balance in data-driven super-resolution. By framing downscaling as a frequency-aware reconstruction problem, our method aligns with the objectives of statistical downscaling while leveraging recent advances in Transformer architectures and implicit neural representations to improve the recovery of physically meaningful fine-scale variability.

\subsection{Relation to Recent Transformer based and Foundation Models}
Recent advances in computer vision have introduced transformer architectures, namely DETR \cite{carion2020end} and Swin Transformer \cite{liu2021swin}, which have proven effective across a wide range of computer vision tasks, including detection, recognition, and semantic segmentation. These models, by their intended application, namely discrete/semantic tasks, focus on the task of feature extraction rather than the task of field reconstruction. Hence, they were not designed to address the relevant computer vision challenges of spectral bias or scale-dependent representation.

In parallel, large-scale vision foundation models are pre-trained on large natural image datasets and have been shown to exhibit unparalleled generalization capabilities \cite{he2022masked,Dosovitskiy2020}. However, their straightforward application to climate super-resolution is limited by specific fundamental differences in data characteristics, learning objectives, and evaluation criteria. Climate fields are continuous physical variables that require strong multiscale, high-fidelity reconstruction, which requires preserving small-scale spatial variability rather than semantic abstraction. In contrast, ViSIR and ViFOR are tailored for dense field reconstruction, explicitly addressing spectral bias through sinusoidal implicit representations and Fourier-based frequency separation guided by Vision Transformer–derived global context. Rather than emphasizing semantic representation, these models focus on continuously reconstructing climate variables at scale, serving as frequency-aware complements to more general-purpose transformer architectures.

\section{Methodology}
\label{sec-method}

\subsection{Problem Formulation}
The super-resolution problem is formulated as learning a mapping function 
\( f_\theta: I_{LR} \rightarrow I_{HR} \), where \( I_{LR} \) is a low-resolution (LR) image and \( I_{HR} \) is the reconstructed high-resolution (HR) output. For Earth System Model (ESM) data, LR inputs are generated by interpolating fine-resolution outputs from the E3SM-HR dataset onto a coarse \(1^\circ \times 1^\circ\) grid. At the same time, HR references are the original
high-resolution (\(0.25^\circ \times 0.25^\circ\)) grid. This resolution mapping corresponds to a $4\times$ super-resolution factor along each spatial dimension. Each image contains normalized variables, including surface temperature, shortwave flux, and longwave flux, on a \(720 \times 1440\) grid.
The \emph{super-resolution} task is then well-suited for the recovery of fine-scale gradients and localized variations
from coarse-resolution ESM outputs.

\begin{figure}[!ht]
 \centering
 \includegraphics[width=\linewidth]{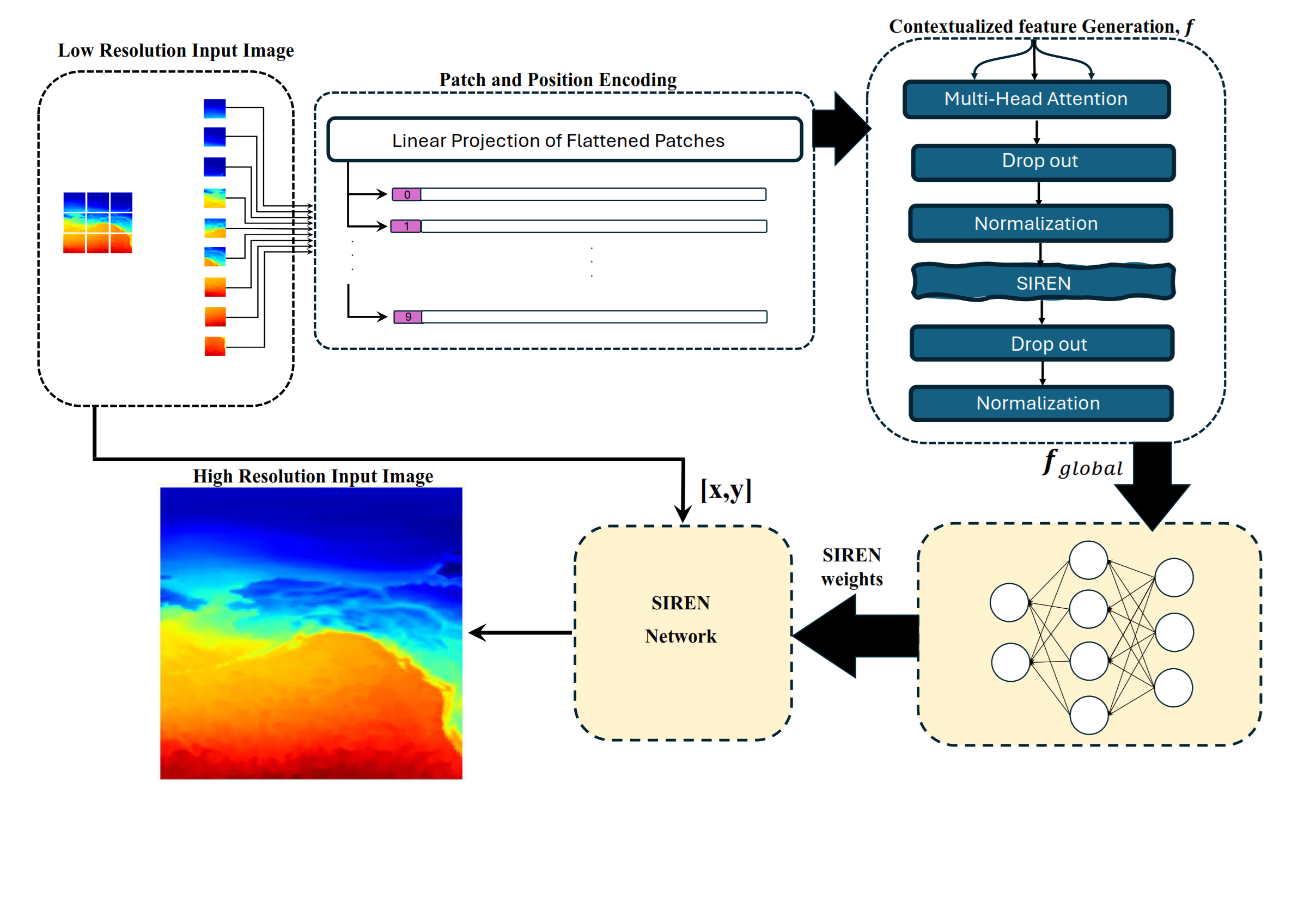} 
 \vspace*{-2em}
 \caption{ViSIR architecture: A low-resolution input image is divided into patches and processed through a Vision Transformer encoder for global context extraction, followed by a hypernetwork-modulated SIREN decoder for frequency-aware continuous reconstruction to extract frequency-aware global features to pass through a hypernetwork. Finally, a coordinate based modulated SIREN decoder maps arbitrary spatial coordinates to RGB values, enabling continuous super-resolution at any target resolution.}
 \label{fig-VITSIREN-architecture}
 \vspace*{-1em}
\end{figure}

\subsection{Vision Transformer-Modulated Sinusoidal Implicit Neural Representation}
\label{VISR}
The Vision Transformer-Modulated Sinusoidal Implicit Neural Representation (ViSIR) is a coordinate based super-resolution framework that combines the global contextual modeling capabilities of Vision Transformers with the frequency-aware properties of implicit neural representations. ViSIR introduces frequency awareness through a coordinate based sinusoidal implicit neural representation (INR) decoder, whose parameters are modulated by global features extracted from a Vision Transformer encoder. Unlike conventional super-resolution methods that operate on discrete pixel grids with fixed upscaling factors, ViTSIREN learns a continuous function that maps spatial coordinates to RGB values, enabling super-resolution reconstruction at arbitrary output resolutions from a single trained model.

Figure~\ref{fig-VITSIREN-architecture} illustrates the overall ViSIR pipeline, which consists of three main components: (i) a Vision Transformer encoder with SIREN based feed-forward networks that extracts frequency-rich global features from the low-resolution input, (ii) a hypernetwork that generates image-specific modulation parameters, and (iii) a modulated SIREN decoder that maps coordinate queries to RGB values conditioned on the extracted features.

\subsubsection{Patch Embedding and Feature Extraction}

Given a low-resolution input image $I_{\text{LR}} \in \mathbb{R}^{H_{\text{LR}} \times W_{\text{LR}} \times 3}$, the image is first partitioned into non-overlapping patches of size $P \times P$. Each patch is linearly embedded into a latent feature space to form a sequence of tokens:

\begin{equation}
E_i = W_p \cdot \text{patch}(I_{\text{LR}})_i + b_p, \quad i = 1,\dots,N,
\label{eq-patch_embedding}
\end{equation}

\noindent where $W_p \in \mathbb{R}^{D \times (P^2 \cdot 3)}$ and $b_p \in \mathbb{R}^D$ are learnable projection parameters, $D$ is the embedding dimension, and $N$ is the total number of patches.

To preserve spatial information, positional encodings are added to the embedded tokens:

\begin{equation}
X_0 = [E_1, E_2, \dots, E_N] + E_{\text{pos}},
\label{eq-pos_encoding}
\end{equation}

\noindent where $E_{\text{pos}} \in \mathbb{R}^{N \times D}$ contains learnable positional embeddings.

The resulting token sequence $X_0$ is then processed through $L$ transformer encoder layers. Each encoder layer consists of multi-head self-attention (MHSA) followed by a SIREN based feed-forward network (FFN):

\begin{equation}
\begin{aligned}
X'_\ell &= X_{\ell-1} + \text{MHSA}(\text{LayerNorm}(X_{\ell-1})), \\
X_\ell &= X'_\ell + \text{FFN}_{\text{SIREN}}(\text{LayerNorm}(X'_\ell)),
\end{aligned}
\label{eq-transformer_layer}
\end{equation}

\noindent where $\ell = 1, \dots, L$ indexes the encoder layers. The multi-head self-attention mechanism enables the model to capture long-range dependencies and global context across the entire low-resolution image.

Unlike conventional Vision Transformers that employ ReLU or GELU activations in the feed-forward network, ViSIR replaces the standard nonlinearity with sinusoidal activations to preserve high-frequency information during feature transformation. The SIREN based FFN is defined as:

\begin{equation}
\text{FFN}_{\text{SIREN}}(x) = \sin\left(\omega_0 \left(W_2 \sin\left(\omega_0 (W_1 x + b_1)\right) + b_2\right)\right),
\label{eq-siren_ffn}
\end{equation}

\noindent where $W_1 \in \mathbb{R}^{D \times D_{\text{FFN}}}$, $W_2 \in \mathbb{R}^{D_{\text{FFN}} \times D}$, $b_1 \in \mathbb{R}^{D_{\text{FFN}}}$, and $b_2 \in \mathbb{R}^D$ are learnable parameters, $D_{\text{FFN}}$ is the FFN hidden dimension (typically $D_{\text{FFN}} = 4D$), and $\omega_0$ is the frequency scaling factor. By introducing sinusoidal nonlinearities at the encoder level, ViSIR preserves high-frequency spatial information throughout the attention based feature aggregation process, mitigating the spectral bias commonly observed in standard transformer architectures.

After processing through all encoder layers, the output tokens are normalized and pooled to obtain a global feature vector that summarizes the image content:

\begin{equation}
f_{\text{global}} = \text{Pool}(\text{LayerNorm}(X_L)) \in \mathbb{R}^D,
\label{eq-global_features}
\end{equation}

\noindent where $\text{Pool}(\cdot)$ denotes an average pooling operation across the spatial dimension. This global feature vector encapsulates the semantic and structural information from the low-resolution input.

\subsubsection{Hypernetwork for Image-Specific Modulation}

To adapt the implicit decoder to image-specific global context while retaining a shared continuous representation across samples, ViSIR employs a lightweight hypernetwork. Rather than using a fixed implicit neural representation for all images, ViSIR employs a hypernetwork to generate image-specific modulation parameters. This design enables the coordinate based decoder to adapt its behavior to the content of each input image.

The hypernetwork is a small multi-layer perceptron that takes the global feature vector $f_{\text{global}}$ as input and produces modulation parameters for each layer of the subsequent SIREN decoder:

\begin{equation}
\theta_{\text{mod}} = \text{MLP}_{\text{hyper}}(f_{\text{global}}),
\label{eq-hypernetwork}
\end{equation}

\noindent where $\theta_{\text{mod}}$ contains scale and shift parameters for modulating the SIREN activations. Specifically, for a SIREN decoder with $M$ hidden layers, the hypernetwork outputs:

\begin{equation}
\theta_{\text{mod}} = \{\gamma_1, \beta_1, \gamma_2, \beta_2, \dots, \gamma_M, \beta_M\},
\label{eq-modulation_params}
\end{equation}

\noindent where $\gamma_m, \beta_m \in \mathbb{R}^{D_{\text{hidden}}}$ are the scale and shift parameters for the $m$-th SIREN layer, and $D_{\text{hidden}}$ is the hidden dimension of the SIREN network.

\subsubsection{Modulated SIREN Decoder}

The core of ViTSIREN's continuous representation is the modulated SIREN decoder, which maps arbitrary spatial coordinates to RGB color values. For any target coordinate $(x, y) \in [-1, 1]^2$ in normalized image space, the decoder computes:

\begin{equation}
\text{RGB}(x, y) = f_{\text{SIREN}}([x, y]; \theta_{\text{mod}}),
\label{eq-coordinate_to_rgb}
\end{equation}

\noindent where $f_{\text{SIREN}}$ is the modulated sinusoidal network.

The modulated SIREN consists of multiple layers with sinusoidal activations. For the first layer:

\begin{equation}
h_1 = \sin\left(\omega_0 \left(W_1 [x, y] + b_1\right)\right),
\label{eq-siren_layer_1}
\end{equation}

\noindent where $W_1 \in \mathbb{R}^{D_{\text{hidden}} \times 2}$, $b_1 \in \mathbb{R}^{D_{\text{hidden}}}$ are learnable parameters, and $\omega_0$ is a frequency scaling factor. The effectiveness of SIREN in mitigating spectral bias stems from the intrinsic periodicity of sinusoidal activation functions. Unlike conventional monotonic activations (e.g., ReLU or GELU), which bias neural networks toward low-frequency representations, sinusoidal activations enable neurons to represent oscillatory patterns at multiple spatial frequencies. In particular, the sinusoidal function forms a Fourier-like basis, allowing linear combinations of sinusoidal responses at different phases and amplitudes to approximate high-frequency components efficiently. The frequency scaling parameter $\omega_0$ directly controls the highest resolvable spatial frequencies, thereby expanding the representational bandwidth of the network. As a result, high-frequency details are not progressively suppressed during training, but instead remain explicitly accessible throughout the feature transformation and reconstruction stages.

For subsequent hidden layers $m = 2, \dots, M$, modulation is applied before the sinusoidal activation:

\begin{equation}
h_m = \sin\left(\omega_0 \left((W_m h_{m-1} + b_m) \odot (1 + \gamma_m) + \beta_m\right)\right),
\label{eq-modulated_siren_layer}
\end{equation}

\noindent where $\odot$ denotes element-wise multiplication. The modulation parameters $\gamma_m$ and $\beta_m$, produced by the hypernetwork, scale and shift the pre-activation values, enabling the network to adapt its frequency response to image content.

Finally, the output layer produces RGB values:

\begin{equation}
\text{RGB}(x, y) = W_{\text{out}} h_M + b_{\text{out}},
\label{eq-siren_output}
\end{equation}

\noindent where $W_{\text{out}} \in \mathbb{R}^{3 \times D_{\text{hidden}}}$ and $b_{\text{out}} \in \mathbb{R}^3$.

\subsubsection{Super-Resolution via Coordinate Querying}

To generate a super-resolved image at target resolution $H_{\text{HR}} \times W_{\text{HR}}$, ViSIR samples coordinates uniformly over the normalized spatial domain $[-1, 1]^2$, and the modulated SIREN decoder processes each coordinate independently to produce the corresponding RGB value:

\begin{equation}
I_{\text{HR}}[i, j] = \text{RGB}(x_i, y_j).
\label{eq-hr_reconstruction}
\end{equation}

Because the decoder operates on continuous coordinates rather than a fixed pixel grid, the same trained model can generate outputs at any desired resolution by simply adjusting the coordinate sampling density. This resolution-agnostic property distinguishes ViSIR from traditional convolutional super-resolution networks that are constrained to specific upscaling factors determined during training.





Algorithm~\ref{alg:vitsiren} summarizes the complete ViSIR training and inference procedure.

\begin{algorithm}[H]
\caption{ViTSIREN: Training and Inference}
\label{alg:vitsiren}
\begin{algorithmic}[1]
\Require Low-resolution image $I_{\text{LR}}$, ground truth $I^*_{\text{HR}}$, patch size $P$, encoder layers $L$, hidden dimension $D$, SIREN layers $M$, frequency parameter $\omega_0$, target resolution $(H_{\text{HR}}, W_{\text{HR}})$
\Ensure Super-resolved output $I_{\text{HR}}$

\State \textbf{// Feature Extraction}
\State Divide $I_{\text{LR}}$ into patches and embed to tokens $X_0 \in \mathbb{R}^{N \times D}$
\State Add positional encodings to $X_0$
\For{$\ell = 1$ to $L$}
    \State $X'_\ell \leftarrow X_{\ell-1} + \text{MHSA}(\text{LN}(X_{\ell-1}))$
    \State $X_\ell \leftarrow X'_\ell + \text{FFN}_{\text{SIREN}}(\text{LN}(X'_\ell))$ \Comment{SIREN based FFN}
\EndFor
\State $f_{\text{global}} \leftarrow \text{Pool}(\text{LN}(X_L))$

\State \textbf{// Modulation Generation}
\State $\theta_{\text{mod}} \leftarrow \text{MLP}_{\text{hyper}}(f_{\text{global}})$
\State Parse $\theta_{\text{mod}}$ into $\{\gamma_1, \beta_1, \dots, \gamma_M, \beta_M\}$

\State \textbf{// Coordinate based Reconstruction}
\State Generate coordinate grid $\mathcal{C}$ at resolution $(H_{\text{HR}}, W_{\text{HR}})$
\For{each $(x, y) \in \mathcal{C}$}
    \State $h_1 \leftarrow \sin(\omega_0 (W_1 [x,y] + b_1))$
    \For{$m = 2$ to $M$}
        \State $h_m \leftarrow \sin(\omega_0 ((W_m h_{m-1} + b_m) \odot (1 + \gamma_m) + \beta_m))$
    \EndFor
    \State $I_{\text{HR}}[x, y] \leftarrow W_{\text{out}} h_M + b_{\text{out}}$
\EndFor

\State \textbf{// Training} (if ground truth available)
\State Compute loss: $\mathcal{L} = \text{MSE}(I_{\text{HR}}, I^*_{\text{HR}})$
\State Update parameters via backpropagation

\State \Return $I_{\text{HR}}$
\end{algorithmic}
\end{algorithm}

\subsection{Motivation for Transition from ViSIR to ViFOR}

ViSIR mitigates spectral bias primarily by employing a sinusoidal implicit decoder guided by global Transformer features, thereby improving the presentation of high-frequency components. This design is effective when the target signal exhibits a reasonably uniform distribution in the frequency domain. However, Earth System Model (ESM) outputs typically comprise multiple physical variables whose spatial fields exhibit markedly different spectral characteristics. For example, surface temperature fields are often dominated by low-wavenumber structures, while radiative fluxes and localized gradients contain stronger high-frequency content.

In ViSIR, a single sinusoidal activation function with a fixed frequency parameter \(\omega_0\) is applied uniformly to all variables and spatial features. While this is a more effective way to model high-frequency content than previous MLP activation functions, it applies a fixed frequency response to all variables. As a result, tuning \( \omega_0 \) involves a trade-off: values that favor high-frequency reconstruction for one variable might be detrimental to the other.

We introduce the ViFOR to address this limitation by replacing the single-frequency sinusoidal activation with an explicit frequency-decomposition mechanism. Using parallel low-pass and high-pass Fourier based activation branches, FOREN and ViFOR separate the latent representation into complementary spectral bands and learn their contributions independently. Conceptually, FOREN shares precisely the same architectural philosophy as SIREN but with a different choice of the activation mechanism: the sinusoidal activations are replaced by frequency-selective Fourier filters. These branches are weighted to fuse, thereby enabling the model to adaptively balance large-scale smooth structures and localized high-frequency details, without relying on any single global frequency parameter.

This explicit separation of frequencies further provides a more flexible and interpretable mechanism for handling multi-variable ESM data, whereby distinct physical processes naturally manifest at different spatial scales. Hence, ViFOR generalizes the frequency-aware design of ViSIR while offering improved robustness and stability across variables with diverse spectral properties.

\subsection{Vision Transformer Fourier Representation Network}
\label{subsec:ViFOR}

\begin{figure}[!ht]
 \centering
 \includegraphics[width=0.9\linewidth]{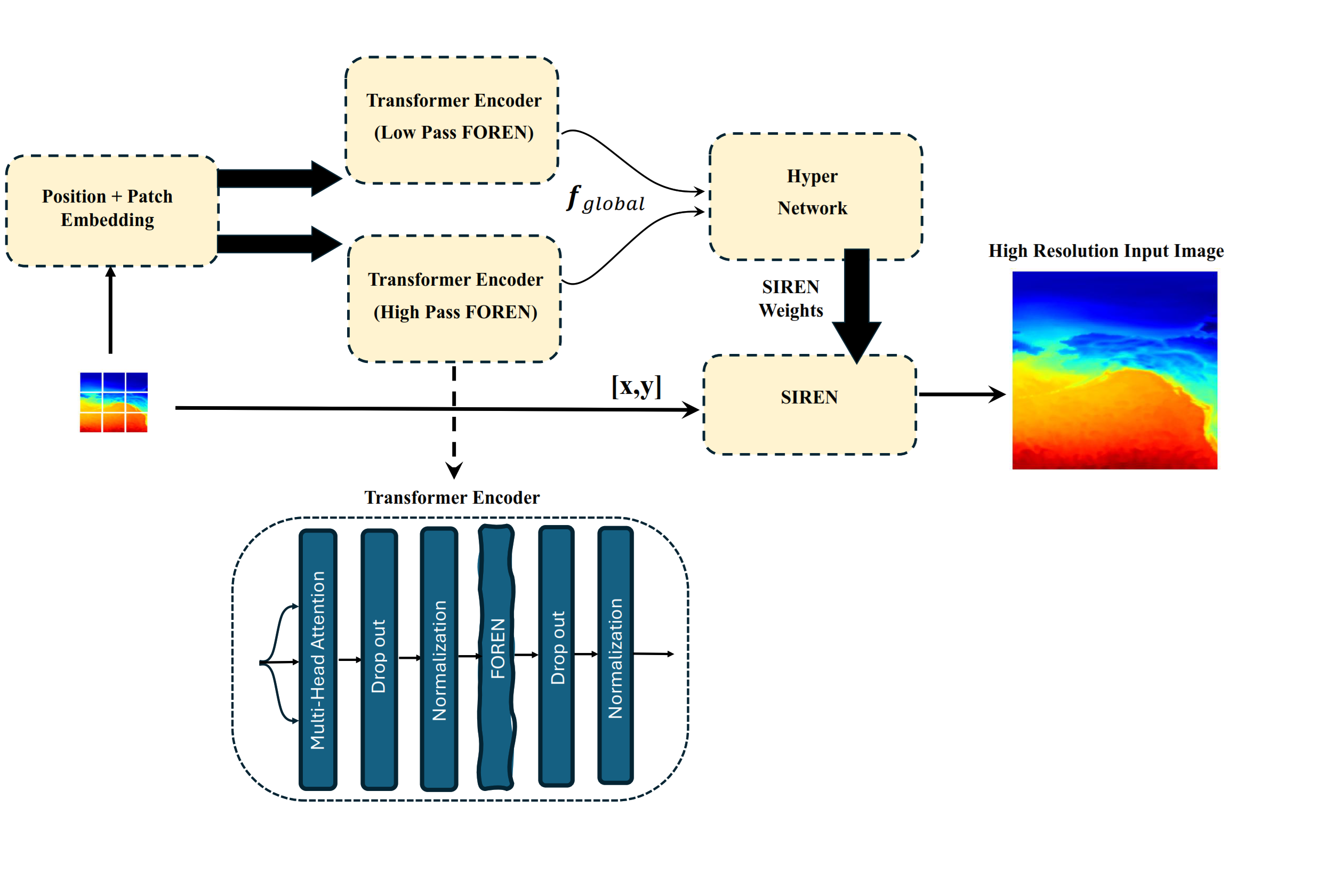} 
 \vspace*{-2em}
 \caption{ViFOR pipeline. In contrast to ViSIR, the sinusoidal feedforward networks are replaced by the Fourier Representation Network (FOREN) within each Transformer encoder block, enabling explicit modulation of low- and high-wavenumber features.}
 \label{fig-ViFOR Flowchart}
 \vspace*{-1em}
\end{figure}

As illustrated in Figure~\ref{fig-ViFOR Flowchart}, the input low-resolution image \( \mathbf{I}_{\text{LR}} \) is first partitioned into non-overlapping patches. Each patch is flattened and embedded using the same linear projection and positional encoding strategy as ViSIR explained in Section \ref{VISR}. Following patch embedding, dropout and layer normalization are applied to stabilize training and normalize the token representations before frequency-aware processing.

The normalized token sequence is then processed by parallel low-pass and high-pass FOREN branches. Each FOREN branch acts as a nonlinear activation module, analogous to the SIREN based feed-forward networks in ViSIR, but operating via the Fourier domain 
filtering. Specifically, each branch applies a Fourier transform to the token features, emphasizes a designated frequency band through an ideal low-pass or high-pass filter, and transforms the filtered features back to the feature domain. This operation serves as a frequency-selective activation function that explicitly separates large-scale smooth structures from localized high-frequency details.

The outputs of the low-frequency and high-frequency FOREN branches are then fused through a weighted summation:

\begin{equation}
\mathbf{Z}_{\text{FOREN}} = \alpha \cdot \mathbf{Z}_{\text{low}} + (1 - \alpha) \cdot \mathbf{Z}_{\text{high}},
\quad \alpha \in [0,1],
\end{equation}

where \( \alpha \) is a learnable parameter controlling the relative contribution of low- and high-wavenumber components. This fusion mechanism plays a role analogous to the frequency-scaling parameter \(\omega_0\) in SIREN, but provides explicit and interpretable control over spectral separation.

The fused frequency-aware representation is subsequently propagated through the Transformer encoder, where attention mechanisms operate on spectrally decomposed features while residual connections and normalization layers remain unchanged. 

Similar to ViSIR, the FOREN based ViT feature vector output is used to train the MLP hypernetwork to provide the scales and shifts required in the FOREN based reconstruction stage, mapping the input LR coordination to the high-resolution output. Using the same Fourier based activation principle at the output stage enables consistent frequency-aware reconstruction and further enhances the recovery of fine-scale spatial details. In contrast to ViSIR, which relies on sinusoidal modulation at both the encoder and reconstruction stages, ViFOR achieves high-fidelity super-resolution through explicit frequency separation using Fourier based activations, resulting in improved stability and robustness across heterogeneous ESM variables.

\subsection{Dataset and Preprocessing}
\begin{figure}[!ht]
\centering
\begin{minipage}{0.45\linewidth}
    \centering
    \includegraphics[width=\linewidth]{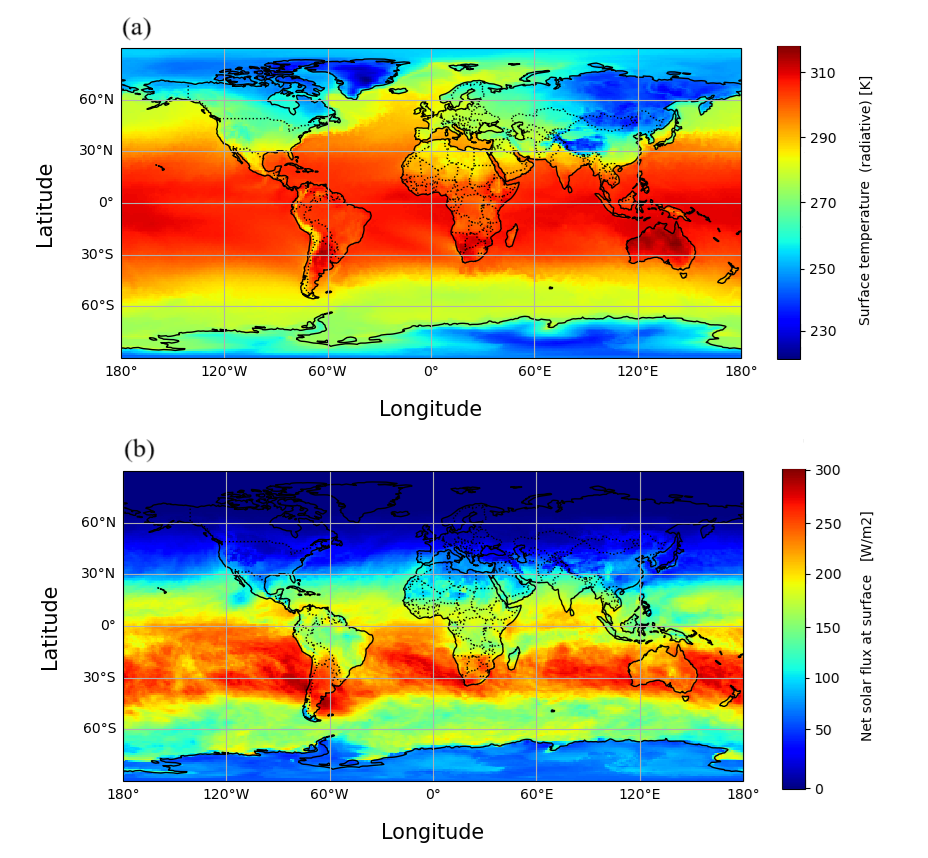}
\end{minipage}
\hfill
\begin{minipage}{0.45\linewidth}
    \centering
    \includegraphics[width=\linewidth]{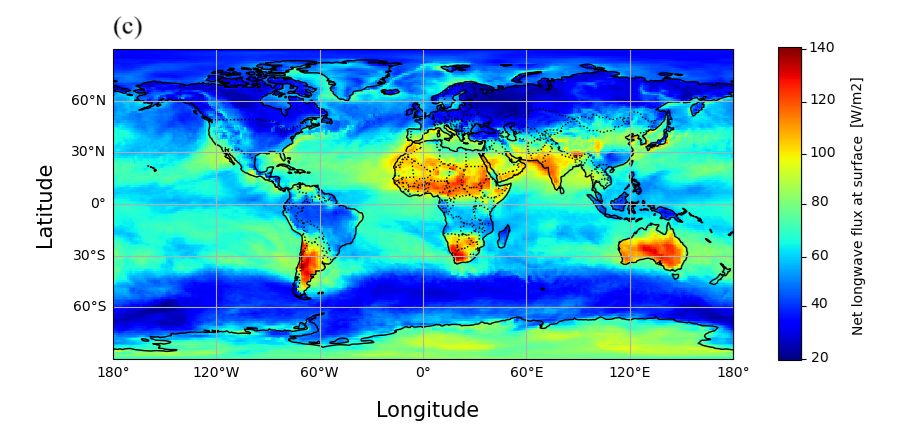}
    \caption{Panels (a), (b), and (c) show surface temperature, shortwave heat flux, and longwave heat flux, respectively, for the first month of year one obtained from the global fine-resolution configuration of E3SM.}
    \label{fig-global}
\end{minipage}
\end{figure}

We trained our models on the \emph{Energy Exascale Earth System Model High Resolution (E3SM-HR)} dataset. The dataset comprises 30 years of monthly control simulations ($30 \times 12$ images). We focused on three significant climate variables shown in Figure~\ref{fig-global}: Surface Temperature (TS), Shortwave Flux (FSW), and Longwave Flux (FLW). All variables were mapped and normalized to RGB channels to create multivariate image representations. For scalability of the model, we employed two settings: (1) \emph{sub-image training}, where an image was divided into non-overlapping patches, and (2) \emph{full-image training}, where global fields were utilized in their entirety. The E3SM simulation
data undergo bilinear interpolation from their original non-orthogonal cubed-sphere grid to a regular $0.25^o\times0.25^o$ longitude-latitude grid, resulting in the interpolated model data known as E3SM-FR \cite{Passarella2022}, which are used only as supervision targets and are not provided as model inputs. To generate the corresponding coarse-resolution input data, the ﬁne-resolution data is further interpolated onto a $1^\circ \times 1^\circ$ grid using a bicubic (BC) method. The dataset in each variable is randomly divided into training, testing, and validation sets with $75\%$, $15 \%$ and $15\%$, respectively. Accordingly, all baselines are trained using each E3SM variable separately, as shown in Table~\ref{tab-Compare1}.


\section{Experimental Setup}
\subsection{Baseline Models}

To comprehensively evaluate the proposed ViSIR and ViFOR frameworks, we compare them against a diverse set of representative super-resolution baselines spanning convolutional, adversarial, transformer based, and implicit-representation paradigms. Specifically, \textbf{SRCNN} serves as a classical CNN based super-resolution model trained end-to-end in the spatial domain \cite{Dong2016}, while \textbf{SRGAN} introduces adversarial learning to enhance perceptual realism in reconstructed outputs \cite{ledig2017}. To represent transformer-based methods, we include a standard \textbf{Vision Transformer (ViT)} super-resolution model employing patch embedding and global self-attention \cite{Dosovitskiy2020}, as well as \textbf{SwinIR}, a state-of-the-art window based transformer that leverages hierarchical shifted-window attention for efficient and scalable image restoration \cite{liu2021swin}. Finally, \textbf{SIREN} is included as a representative implicit neural representation method that employs sinusoidal activations to mitigate spectral bias in continuous signal reconstruction \cite{SIREN}. Together, these baselines ensure a balanced and rigorous comparison across spatial-domain, frequency-aware, transformer based, and implicit-representation super-resolution methods, enabling a fair assessment of the advantages offered by ViSIR and ViFOR.



\subsection{Evaluation Metrics}
To evaluate the algorithm's performance, we employ three complementary metrics. Mean Squared Error (MSE) to quantify the average pixel-wise difference between the reconstructed HR image and the ground truth, Peak Signal-to-Noise Ratio (PSNR) to measure reconstruction quality in decibels, where higher values indicate less distortion, and Structural Similarity Index (SSIM) to assess perceived image similarity, emphasizing structural fidelity\cite{SSIM}.

These metrics capture both numerical accuracy (MSE, PSNR) and perceptual quality (SSIM), which are significant for assessing ESM data reconstructions.  

All methods are trained using a standard pixel-wise mean squared error (MSE) loss between the predicted high-resolution output and the ground-truth field:
\begin{equation}
\mathcal{L}_{\mathrm{MSE}} = \frac{1}{N} \sum_{i=1}^{N} 
\left\| \hat{\mathbf{I}}_{\mathrm{HR}}^{(i)} - \mathbf{I}_{\mathrm{HR}}^{(i)} \right\|_2^2 
\end{equation}

No explicit frequency-domain loss is imposed during the training of ViSIR and ViFOR. Instead, enhanced high-frequency recovery arises implicitly from the frequency-aware architectural components, namely the sinusoidal activations in ViSIR and the Fourier based FOREN modules in ViFOR.

\subsection{Training Environment and Hyperparameters}
\label{hyperparameter}

Both ViSIR and ViFOR were trained on the E3SM-HR dataset, which includes 360 monthly images spanning thirty years across three variables: surface temperature, shortwave flux, and longwave flux. All ML models in Table~\ref{tab-Compare1} are applied to each variable independently and were optimized using the Adam optimizer with an initial learning rate of \(10^{-4}\) and a cosine decay schedule. Mean squared error was used as the primary loss function, with additional perceptual loss experiments conducted to evaluate qualitative improvements.  

\begin{figure}[!ht]
\centering
\begin{minipage}{0.45\linewidth}
    \centering
    \includegraphics[width=\linewidth]{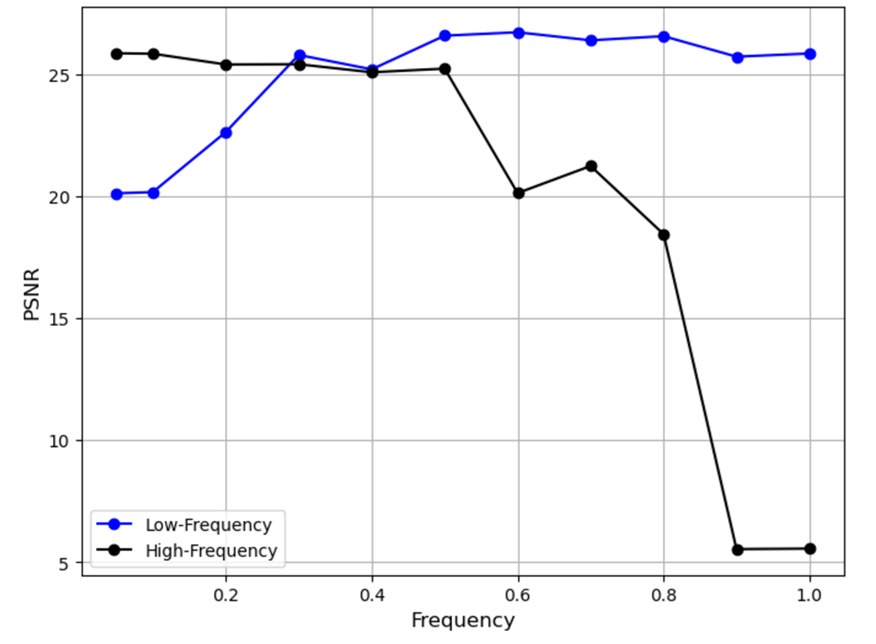}
\end{minipage}
\hfill
\begin{minipage}{0.35\linewidth}
    \caption{PSNR across different cutoff frequencies $f_c$ for ViFOR. Optimal performance was achieved at $f_c=0.3$.}
    \label{fig-cutoff}
\end{minipage}
\end{figure}

In addition to optimization settings, ViSIR and ViFOR include several architecture-specific hyperparameters that control frequency behavior and model capacity. For the Vision Transformer backbone used in both models, we employ a patch size of $16 \times 16$, an embedding dimension of 256, 6 Transformer encoder layers, and 8 attention heads (each operating on 32 dimensions) and feed-forward networks with hidden dimension $D_{\text{FFN}} = 1024$. The SIREN decoder comprises 5 layers, while the hypernetwork uses 2 hidden layers of 512 nodes each to generate 2048 modulation parameters (scale and shift values) for the SIREN layers, enabling image-specific frequency adaptation. These architectural choices balance model expressiveness with computational efficiency, allowing both ViSIR and ViFOR to capture long-range spatial dependencies while maintaining reasonable training times on standard hardware.

\begin{figure}[!ht] 
 \centering
 \includegraphics[width=\linewidth]{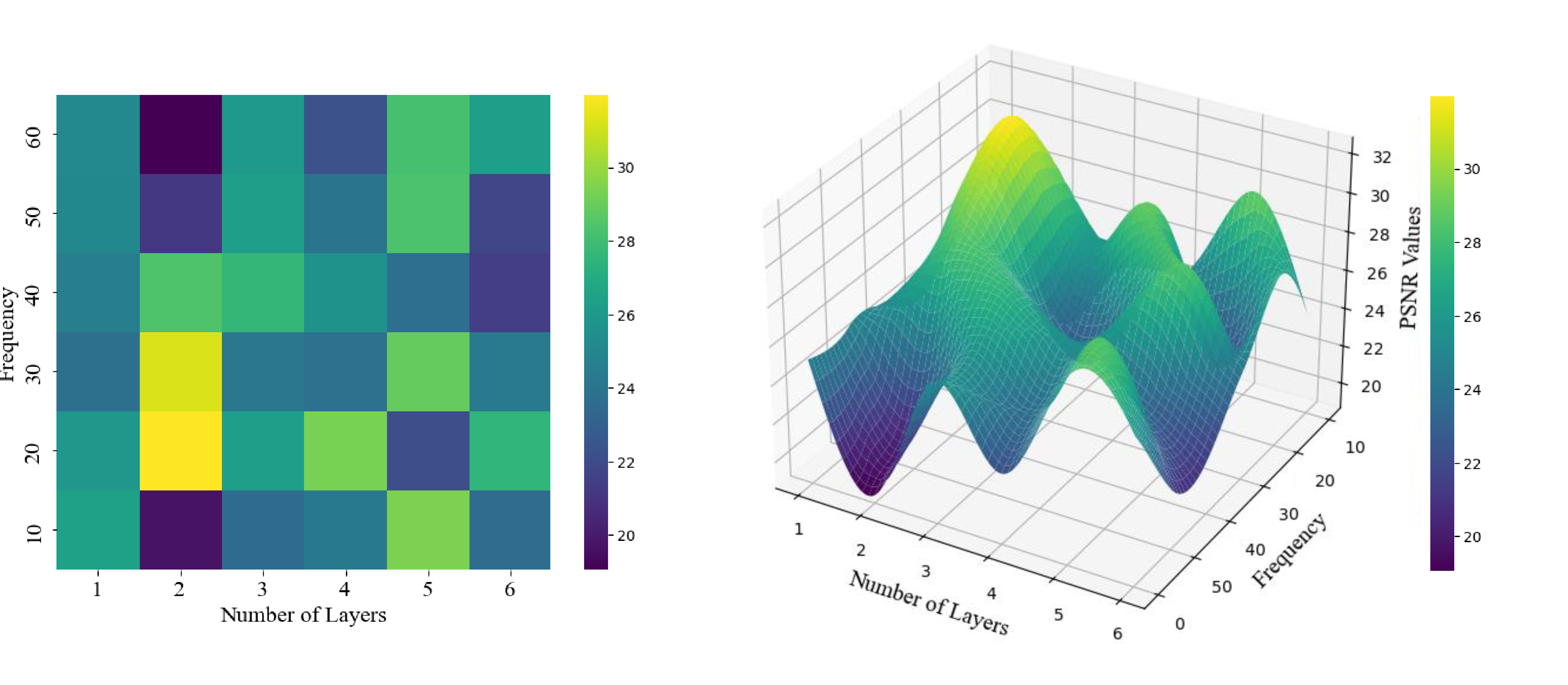} 
 \caption{2D (left) and 3D (right) illustration of the PSNR values for different Frequencies and different numbers of hidden layers used in the proposed ViSIR applied to the E3SM dataset }
 \label{fig-FreqVsLayer} 
\end{figure}

For ViFOR, cutoff frequencies, defined as spatial-frequency components of two-dimensional image fields, were investigated between 0.01 and 1, with 0.3 identified as the optimal cutoff for both low-pass and high-pass filters. Cutoff values (e.g., 0.3) are normalized spatial-frequency thresholds in the image spectrum, rather than physical temporal frequencies measured in hertz. They are chosen based on the mean PSNR across all variables. For ViSIR, we investigated PSNR across sinusoidal activation frequencies from 10 to 60 while varying the number of hidden layers in SIREN from 1 to 6. Figure~\ref{fig-FreqVsLayer} illustrates the effect of different parameter combinations. Based on the mean PSNR values for all variables shown in Figure~\ref{fig-FreqVsLayer}, the best PSNR for EMS images is achieved with two hidden layers at a frequency of 20. To ensure a fair comparison, these hyperparameters are used for the remaining methods across frequency and layers.

\section{Experimental Results}
\label{sec-results}

\subsection{Baseline Configuration and Fairness of Comparison}

To ensure a controlled and consistent evaluation, all baseline models and the proposed ViSIR and ViFOR methods were trained and evaluated under the same experimental conditions.

Each model was trained for the same number of epochs using the same optimization strategy and learning-rate schedule, and was evaluated using identical performance 
metrics. Where applicable, baseline architectures were configured to have comparable model capacity without altering their standard, commonly adopted implementations, to avoid disadvantaging any method. This setup helps attribute performance differences primarily to architectural design choices rather than differences in data usage, training strategy, or evaluation protocol. The comparison of all models in terms of architecture, frequency modeling strategy, training objective, and computational cost are shown in Table~\ref{tab-baseline_comparison}.

\begin{table*}[!ht]
\centering
\caption{Comparison of representative super-resolution models in terms of architecture, frequency modeling strategy, training objective, and computational cost.}
\label{tab-baseline_comparison}
\scriptsize
\resizebox{\textwidth}{!}{%
\begin{tabular}{lccccc}
\toprule
\textbf{Model} 
& \textbf{Category} 
& \textbf{Training Objective} 
& \textbf{Params (M)} 
& \textbf{Optimizer}
&\textbf{Speed(itr/s)}\\
\midrule
SRCNN     
& CNN        
& MSE           
& $\sim$0.1 
& Adam
& 11.64\\

SRGAN     
& GAN        
& MSE + Adversarial
& $\sim$1.5 
& Adam
& 6.9\\

ViT-SR    
& Transformer
& MSE           
& $\sim$7.4 
& Adam  
& 2.36\\

SIREN     
& INR        
& MSE           
& $\sim$2.6 
& Adam  
& 11.25\\

SwinIR     
& Transformer        
& MSE           
& $\sim$3.8 
& Adam  
& 4.58\\

\midrule
ViSIR     
& ViT + INR  
& MSE           
& $\sim$4.9 
& Adam 
& 12.8\\

ViFOR     
& ViT + INR  
& MSE           
& $\sim$4.9 
& Adam  
& 11.5\\
\bottomrule
\end{tabular}}
\end{table*}

\subsection{Quantitative Comparison}
Table~\ref{tab-Compare1} summarizes the performance across three ESM variables separately. Each ESM variable is treated independently, and all models are trained and evaluated on the same variable in a single-variable setting to ensure a fair and controlled comparison. The quantitative results in Table~\ref{tab-Compare1} highlight the reconstruction capacity of the novel ViSIR and ViFOR frameworks compared to baselines. Conventional CNN-based methods, such as SRCNN and SRGAN, the transformer-based ViT, and the implicit SIREN models generally achieve peak signal-to-noise ratio (SNR) results below 24–27 dB and structural similarity index (SSIM) values around 0.6–0.7. These findings demonstrate limited capability to reconstruct high-frequency spatial details important for Earth System Model variables.
Nevertheless, ViSIR features a noticeable performance improvement over all the baseline models by mitigating spectral bias through a sinusoidal implicit representation guided by Transformer-extracted global context with PSNR improvements of approximately 5~dB over ViT and subtle superiority over SwinIR, with a more than 50\% reduction of mean squared error of all the variables considered. The improved ViFOR model extends these strengths by incorporating Fourier based activation filtering and transformer attention to enable balanced learning of low- and high-frequency components. ViFOR achieves 29.2~dB, 29.3~dB, and 30.9~dB for Source Temperature, Shortwave Flux, and Longwave Flux, respectively, and SSIM up to 0.80 and MSE values methoding 0.1\%. ViFOR consistently outperforms ViSIR and all other models, with reconstructed ESM fields exhibiting greater fidelity and structural integrity, and with physically relevant spatial gradients and variability intact.

A majority of existing super-resolution methods for climate data are based on convolutional networks or GAN architectures that focus on grid space modeling \cite{vandal2017deepsd,stengel2020adversarial}. These methods proved successful in some instances, but do not account for spectral biases in deep networks when modeling large-scale patterns. Baseline methods such as ViT, which are based on the transformer architecture, are better at modeling global patterns but are not suitable for modeling fine-scale subgrid variations.

On the other hand, ViSIR and ViFOR integrate mechanisms based on frequency awareness into the Transformer architecture. ViSIR employs implicit sinusoidal representations to improve small-scale reconstruction, and ViFOR further extends this strategy to explicitly separate large- and small-scale spatial features via Fourier-activated functions. This enables better reconstruction of multiscale fields without resorting to additional physical variables.

\begin{table*}[!ht]
\centering
\caption{Mean $\pm$ standard deviation of MSE (\%), PSNR (dB), and SSIM [0--1]
computed over five independent runs for original $I_O$ and reconstructed $I_R$
images. Results are reported for Surface Temperature, Shortwave Heat Flux, and
Longwave Heat Flux under sub-image and full-image training settings.}
\renewcommand{\arraystretch}{0.8}
\resizebox{\textwidth}{!}{%
\begin{tabular}{lcccccc}
\toprule
\textbf{} & \multicolumn{3}{c}{\bf Sub-Image} & \multicolumn{3}{c}{\bf Full Image}\\
\textbf{Models} & \textbf{MSE \%} & \textbf{PSNR} & \textbf{SSIM}
& \textbf{MSE \%} & \textbf{PSNR} & \textbf{SSIM} \\
\midrule
\multicolumn{7}{c}{\bf Surface Temperature} \\
\cmidrule(lr){1-7}
ViT & $0.44 \pm 0.01$ & $22.31 \pm 0.08$ & $0.61 \pm 0.006$
    & $0.51 \pm 0.01$ & $23.10 \pm 0.07$ & $0.63 \pm 0.006$ \\
SRCNN & $0.15 \pm 0.03$ & $27.38 \pm 0.18$ & $0.73 \pm 0.014$
       & $0.15 \pm 0.03$ & $27.29 \pm 0.17$ & $0.73 \pm 0.014$ \\
SRGAN & $0.43 \pm 0.04$ & $23.55 \pm 0.22$ & $0.65 \pm 0.018$
       & $0.63 \pm 0.05$ & $21.31 \pm 0.21$ & $0.59 \pm 0.019$ \\
SIREN & $1.20 \pm 0.07$ & $19.18 \pm 0.26$ & $0.51 \pm 0.025$
      & $0.95 \pm 0.06$ & $20.06 \pm 0.24$ & $0.56 \pm 0.024$ \\
SwinIR & $0.14 \pm 0.02$ & $28.43 \pm 0.16$ & $0.75 \pm 0.01$
      & $0.14 \pm 0.03$ & $28.42 \pm 0.20$ & $0.76 \pm 0.008$ \\
      
\textbf{ViSIR} & $0.14 \pm 0.02$ & $28.42 \pm 0.11$ & $0.75 \pm 0.009$
               & $0.14 \pm 0.02$ & $28.33 \pm 0.10$ & $0.75 \pm 0.009$ \\
\textbf{ViFOR} & $0.13 \pm 0.03$ & $28.55 \pm 0.13$ & $0.76 \pm 0.011$
               & $\mathbf{0.12 \pm 0.03}$ & $\mathbf{29.21 \pm 0.14}$ & $\mathbf{0.77 \pm 0.012}$ \\
\midrule
\multicolumn{7}{c}{\bf Shortwave Heat Flux} \\
\cmidrule(lr){1-7}
ViT & $0.49 \pm 0.01$ & $23.00 \pm 0.08$ & $0.63 \pm 0.006$
    & $0.42 \pm 0.01$ & $23.80 \pm 0.07$ & $0.65 \pm 0.006$ \\
SRCNN & $0.18 \pm 0.03$ & $26.98 \pm 0.17$ & $0.72 \pm 0.014$
       & $0.21 \pm 0.03$ & $26.74 \pm 0.16$ & $0.71 \pm 0.014$ \\
SRGAN & $0.43 \pm 0.04$ & $22.67 \pm 0.22$ & $0.64 \pm 0.018$
       & $0.69 \pm 0.05$ & $21.10 \pm 0.21$ & $0.60 \pm 0.019$ \\
SIREN & $0.96 \pm 0.07$ & $19.98 \pm 0.25$ & $0.55 \pm 0.024$
      & $0.93 \pm 0.06$ & $20.41 \pm 0.24$ & $0.57 \pm 0.023$ \\
SwinIR & $0.17 \pm 0.03$ & $27.92 \pm 0.10$ & $0.73 \pm 0.012$
               & $0.14 \pm 0.03$ & $28.12 \pm 0.10$ & $0.74 \pm 0.023$ \\

\textbf{ViSIR} & $0.15 \pm 0.02$ & $28.15 \pm 0.11$ & $0.74 \pm 0.009$
               & $0.16 \pm 0.02$ & $27.88 \pm 0.10$ & $0.73 \pm 0.009$ \\
\textbf{ViFOR} & $0.14 \pm 0.03$ & $28.48 \pm 0.13$ & $0.76 \pm 0.011$
               & $\mathbf{0.11 \pm 0.03}$ & $\mathbf{29.34 \pm 0.14}$ & $\mathbf{0.78 \pm 0.012}$ \\
\midrule
\multicolumn{7}{c}{\bf Longwave Heat Flux} \\
\cmidrule(lr){1-7}
ViT & $0.95 \pm 0.01$ & $20.00 \pm 0.08$ & $0.56 \pm 0.006$
    & $0.93 \pm 0.01$ & $20.40 \pm 0.07$ & $0.57 \pm 0.006$ \\
SRCNN & $0.23 \pm 0.03$ & $26.25 \pm 0.17$ & $0.70 \pm 0.014$
       & $0.24 \pm 0.03$ & $26.02 \pm 0.16$ & $0.70 \pm 0.014$ \\
SRGAN & $0.66 \pm 0.04$ & $21.18 \pm 0.22$ & $0.59 \pm 0.018$
       & $0.93 \pm 0.05$ & $20.39 \pm 0.21$ & $0.57 \pm 0.019$ \\
SIREN & $0.92 \pm 0.07$ & $20.55 \pm 0.25$ & $0.58 \pm 0.024$
      & $0.69 \pm 0.06$ & $21.05 \pm 0.24$ & $0.59 \pm 0.023$ \\
SwinIR & $0.15 \pm 0.02$ & $28.02 \pm 0.08$ & $0.74 \pm 0.009$
               & $0.21 \pm 0.14$ & $26.87 \pm 0.10$ & $0.71 \pm 0.014$ \\

\textbf{ViSIR} & $0.16 \pm 0.02$ & $27.42 \pm 0.11$ & $0.72 \pm 0.009$
               & $0.09 \pm 0.02$ & $30.28 \pm 0.10$ & $0.80 \pm 0.009$ \\
\textbf{ViFOR} & $0.15 \pm 0.03$ & $28.18 \pm 0.13$ & $0.75 \pm 0.011$
               & $\mathbf{0.10 \pm 0.03}$ & $\mathbf{30.96 \pm 0.14}$ & $\mathbf{0.79 \pm 0.012}$ \\
\bottomrule
\end{tabular}}
\label{tab-Compare1}
\end{table*}

\subsection{Frequency-Domain Evaluation and Spectral Fidelity}

\begin{figure}[!ht]
\centering
\begin{subfigure}{\textwidth}
\includegraphics[width=\textwidth]{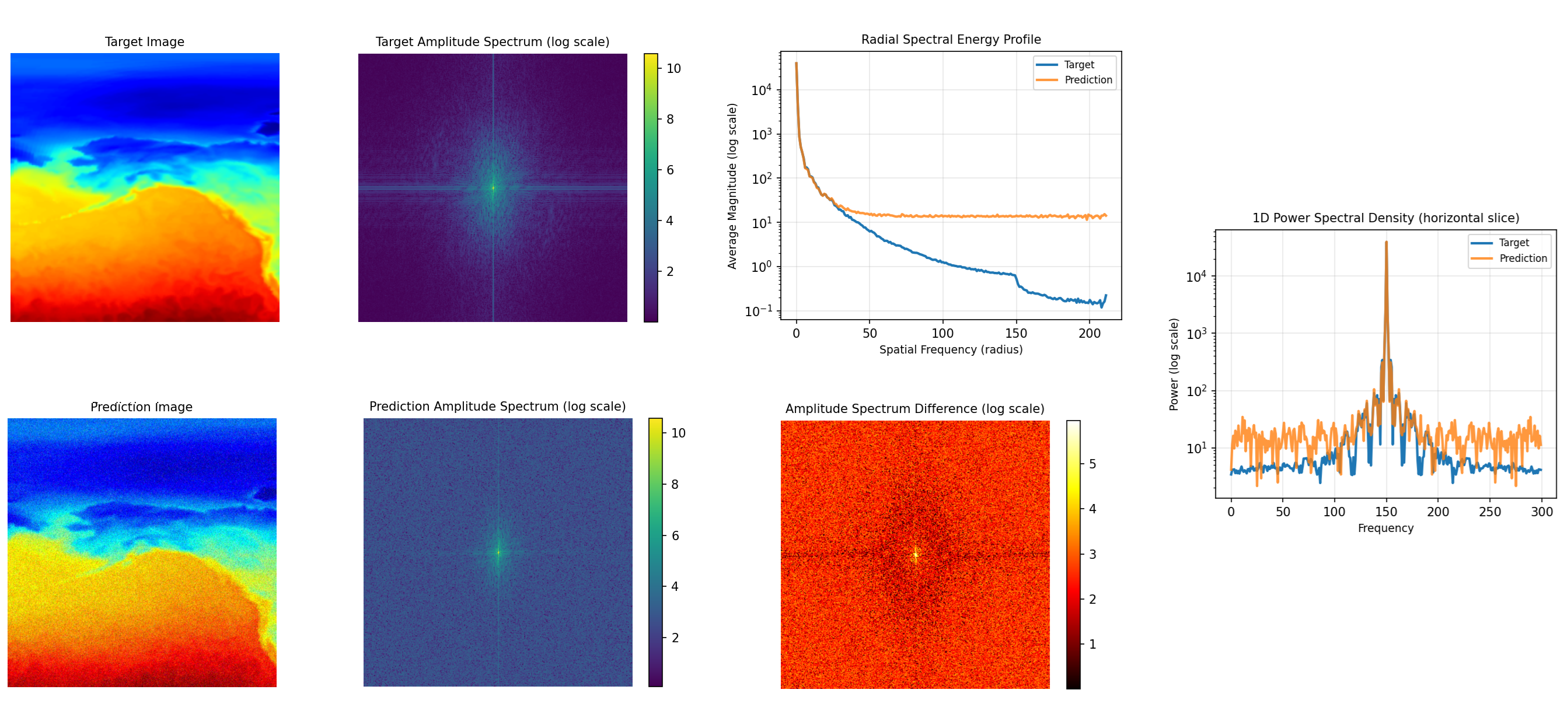}
\caption{ViSIR: Frequency-domain errors in mid-to-high frequency bands. Radial spectral energy profile diverges beyond $f=35$, and 1D PSD shows excessively high-frequency noise.}
\label{fig-ViSIR-frequency}
\end{subfigure}
\vspace{0.3cm}
\begin{subfigure}{\textwidth}
\includegraphics[width=\textwidth]{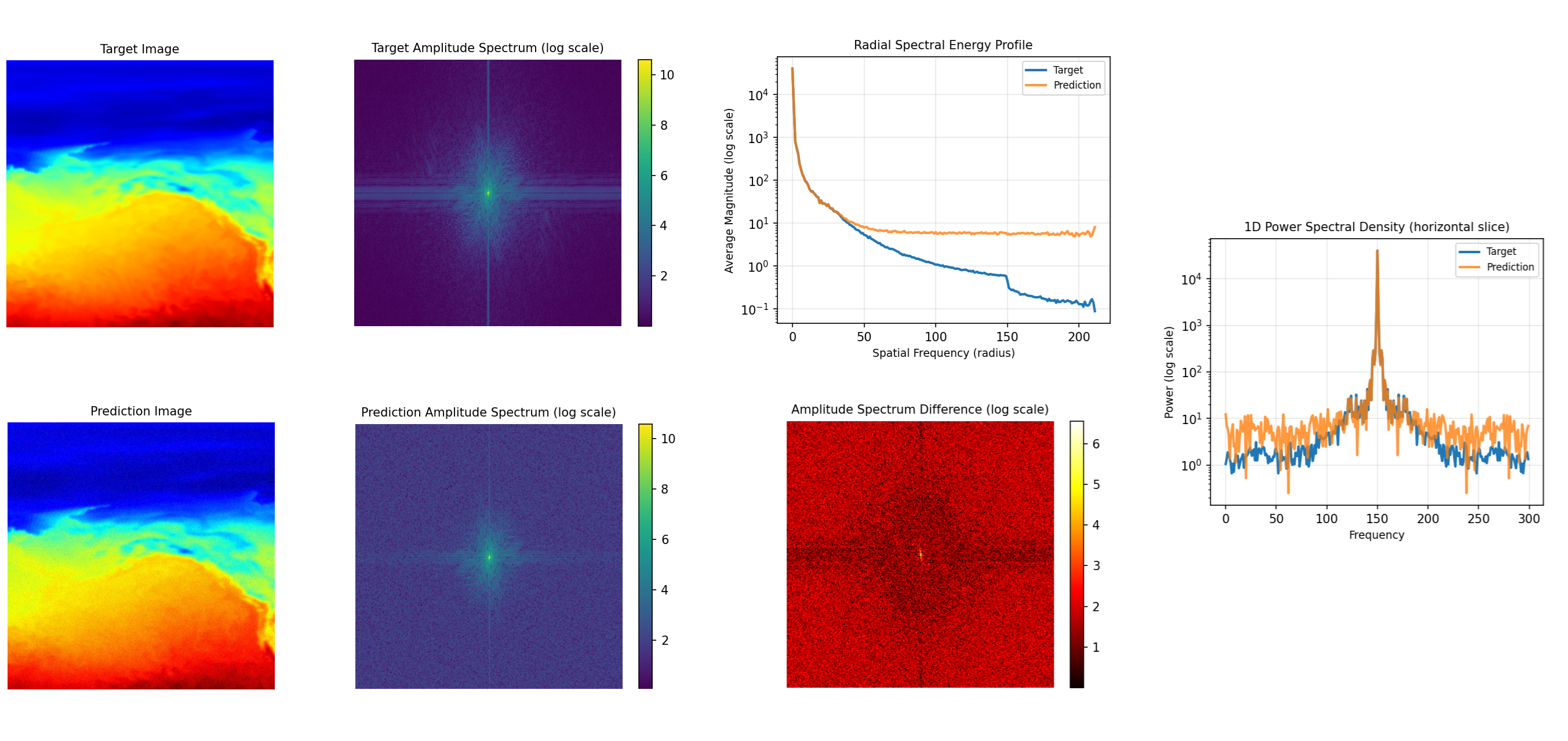}
\caption{ViFOR substantially mitigates spectral bias by recovering mid- and high-frequency components. Radial spectral energy profile diverges beyond $f=50$}
\label{fig-ViFOR-frequency}
\end{subfigure}
\caption{Comparison of the frequency behavior of ViSIR and ViFOR}
\label{fig-frequency_analysis}
\end{figure}

In this section, we present a comprehensive spectral-domain evaluation of the proposed ViSIR and ViFOR models. Figure~\ref{fig-frequency_analysis} visualizes multiple complementary frequency diagnostics, including the 2D Fourier amplitude spectrum, amplitude spectrum error, radial spectral energy profile, and one-dimensional power spectral density (PSD). All frequency indices reported in spectral plots correspond to discrete Fourier wavenumber indices in the spatial frequency domain, while cutoff frequencies $f_c$ denote normalized thresholds relative to the Nyquist frequency. The log-scaled 2D Fourier amplitude spectra reveal how spectral energy is distributed across spatial frequencies. For ViSIR (Figure~\ref{fig-ViSIR-frequency}), the predicted spectrum essentially captures the dominant low-frequency structure of the target field. However, mild attenuation and spatially non-uniform deviations remain visible in the mid-to-high frequency bands. In contrast, ViFOR (Figure~\ref{fig-ViFOR-frequency}) exhibits a markedly closer alignment with the target amplitude spectrum across the entire frequency plane, preserving both the low-frequency energy concentration and the anisotropic high-frequency patterns associated with sharp gradients and localized variability.

Furthermore, considering the amplitude-spectrum difference maps, ViSIR shows localized spectral discrepancies, particularly in intermediate-frequency regions, reflecting the limitations of a single global sinusoidal activation in uniformly balancing frequency content. ViFOR, on the other hand, demonstrates consistently lower spectral error over all frequencies, confirming the effectiveness of explicit low-pass and high-pass Fourier based modulation.

Moreover, radial spectral energy analysis shows that while ViSIR matches the target spectrum at low spatial frequencies but diverges at higher bands, ViFOR maintains close agreement across the full frequency range, demonstrating that explicit Fourier based decomposition more effectively balances large- and small-scale spatial reconstruction and mitigates spectral bias in heterogeneous ESM fields.

The 1D PSD obtained from horizontal frequency slices provides additional insight into directional frequency behavior. ViSIR reproduces the dominant spectral peak and general decay trend but exhibits elevated noise levels and fluctuations at higher frequencies. In contrast, ViFOR closely matches both the peak location and the spectral roll-off of the target PSD, while maintaining a lower high-frequency noise floor. This indicates that ViFOR not only recovers high-frequency content but does so in a controlled and physically consistent manner.


\subsection{Comparison of Sub-image vs. Full-image Training}

To investigate the impact of spatial context on super-resolution performance, we consider two training strategies. In the \emph{sub-image training} setting, each global low-resolution field is partitioned into non-overlapping sub-images. These sub-images correspond to large regional domains and are treated as independent training samples, increasing the effective number of training instances while limiting global spatial context.

In contrast, the \emph{full-image training} setting uses the entire global field as a single input sample, preserving complete spatial coverage and long-range dependencies across the globe. In both settings, the corresponding high-resolution fields at $0.25^\circ \times 0.25^\circ$ resolution are used solely as supervision targets. They are not provided as model inputs to ensure a fair comparison between the two strategies.

The results indicate that both ViSIR and ViFOR benefit from exposure to the full-image domain, with ViFOR exhibiting a more pronounced improvement. In particular, ViFOR achieved 29.2~dB PSNR and 0.77~SSIM on full global images compared to 28.5~dB and 0.76~SSIM on sub-image training for the Surface Temperature variable. This substantial gain underscores the importance of retaining large-scale spatial dependencies inherent in Earth System Model (ESM) data. Owing to its transformer based backbone and Fourier-modulated representations, ViFOR effectively exploits global contextual information, enabling the model to reconstruct high-frequency patterns that are spatially coherent with large-scale climate structures. In contrast, ViSIR exhibits moderate but consistent improvement under full-image training, reflecting its enhanced ability to model local variations through sinusoidal activations, albeit with less global context awareness than ViFOR.\\

\subsection{Ablation Studies}
\paragraph{Effect of Sinusoidal Frequency Parameter in ViSIR}
We examined the influence of the SIREN frequency parameter $\omega_0$. Larger values of $\omega_0$ improved recovery of high-frequency details but at the cost of training instability. An intermediate selection ($\omega_0 = 20$) proved optimal, as mentioned in Section \ref{hyperparameter}. This behavior reflects the trade-off between representational bandwidth and training stability inherent in sinusoidal activations.

\paragraph{Effect of Fourier Cutoff Frequency in ViFOR}
For ViFOR, cutoff frequencies $f_c$ between $0.01$ and $1.0$ were tested. Figure~\ref{fig-cutoff} shows that $f_c=0.3$ offered maximum PSNR on both the low-pass and high-pass branches. Cutoffs above $0.3$ degraded performance by amplifying noise, whereas lower cutoffs limited the recovery of high-frequency information. This confirms that explicit frequency separation provides a more stable alternative to single frequency tuning.

\paragraph{Effect of Input Resolution on Computational Cost}

We analyze how the computational complexity of ViTSIREN scales with input resolution. Table~\ref{tab-resolution-scaling} summarizes the relationship between input size and computational requirements for $3\times$ super-resolution.

\begin{table}[t]
\centering
\caption{Resolution Scaling Analysis for ViTSIREN ($3\times$ Super-Resolution)}
\scriptsize
\label{tab-resolution-scaling}
\begin{tabular}{lccccc}
\toprule
\textbf{Input} & \textbf{Training} & \textbf{Patches} & \textbf{Params} & \textbf{Attention} & \textbf{Total} \\
\textbf{Size} & \textbf{(iter/s)} & $(N)$ & \textbf{(M)} & \textbf{Cost} & \textbf{FLOPs} \\
\midrule
$64 \times 64$ & 24.70 & 64 & 4.89 & 1.00$\times$ & 14.70G \\
$100 \times 100$ & 12.01 & 100 & 4.91 & 2.45$\times$ & 36.10G \\
$128 \times 128$ & 7.53 & 169 & 4.93 & 4.01$\times$ & 58.99G \\
$200 \times 200$ & 3.10 & 400 & 5.10 & 9.81$\times$ & 144.21G \\
\bottomrule
\end{tabular}
\vspace{0.3em}

\footnotesize{Patch size is fixed at $10 \times 10$. Attention cost is reported relative to the $64 \times 64$ baseline.}
\end{table}

The number of model parameters remains nearly constant across resolutions, increasing by only 4.3\% from $64 \times 64$ to $200 \times 200$ inputs. This stability arises because only the positional embeddings scale with input size, while all other components remain fixed. In contrast, computational cost varies significantly across components:

\begin{itemize}
    \item \textbf{Transformer Attention:} The self-attention mechanism scales quadratically with the number of patches, following $\mathcal{O}(N^2)$ complexity where $N = HW/P^2$ for input dimensions $H \times W$ and patch size $P$. As shown in Table~\ref{tab-resolution-scaling}, attention cost increases by $9.81\times$ when moving from $64 \times 64$ to $200 \times 200$ inputs.
    
    \item \textbf{SIREN Decoder:} The implicit neural representation scales linearly with the number of output pixels, i.e., $\mathcal{O}(H_{\text{out}} \times W_{\text{out}})$. For $3\times$ upscaling, each doubling of input resolution results in a $4\times$ increase in decoder computation.
    
    \item \textbf{Hypernetwork:} The modulation parameter generator maintains constant cost regardless of resolution, as it produces a fixed-size output for each input image.
\end{itemize}

Table~\ref{tab-component-breakdown} presents the distribution of computational cost across model components at different resolutions.

\begin{table}[t]
\centering
\caption{Component-wise FLOPs Distribution at Different Input Resolutions}
\scriptsize
\label{tab-component-breakdown}
\begin{tabular}{lcccc}
\toprule
\textbf{Component} & $\mathbf{64^2}$ & $\mathbf{100^2}$ & $\mathbf{200^2}$ & $\mathbf{256^2}$ \\
\midrule
Patch Embedding & 0.2\% & 0.1\% & $<$0.1\% & $<$0.1\% \\
Transformer Encoder & 3.1\% & 5.4\% & 15.8\% & 24.6\% \\
Hypernetwork & $<$0.1\% & $<$0.1\% & $<$0.1\% & $<$0.1\% \\
SIREN Decoder & 96.7\% & 94.5\% & 84.1\% & 75.3\% \\
\bottomrule
\end{tabular}
\end{table}

The SIREN decoder dominates the computational cost across all tested resolutions, accounting for 75.3\%-96.7\% of total FLOPs. However, its relative contribution decreases at higher resolutions due to the quadratic scaling of transformer attention. The encoder's share increases from 3.1\% at $64 \times 64$ to 24.6\% at $256 \times 256$, indicating that attention will become the computational bottleneck for sufficiently large inputs.

These findings have practical implications for deployment. At moderate resolutions ($\leq 200 \times 200$), the linear scaling of the SIREN decoder determines overall efficiency. For higher-resolution applications, incorporating efficient attention mechanisms, such as windowed attention~\cite{liu2021swin} or linear attention approximations, may be beneficial for mitigating quadratic scaling.

\section{Discussion}
\label{sec-discussion}

This study investigated frequency-aware super-resolution for Earth System Model (ESM) data using two Transformer based architectures, ViSIR and ViFOR. The results demonstrate that incorporating sinusoidal and Fourier-based activation functions within Transformer-guided implicit and Fourier-based representation frameworks significantly improves the recovery of small-scale spatial details compared to conventional CNN-, GAN-, and vanilla Transformer-based methods. ViSIR reduces spectral bias primarily by enhancing mid-frequency representations and partially improving high-wavenumber recovery. In contrast, ViFOR further improves reconstruction accuracy by explicitly separating large- and small-scale spatial components. Quantitative improvements in PSNR, SSIM, and MSE, together with frequency-domain diagnostics, confirm that ViFOR achieves a more balanced reconstruction across spatial scales while preserving global coherence.

Despite these gains, several limitations remain. The proposed models are trained in a supervised setting using paired low- and high-resolution data and focus exclusively on spatial super-resolution. Temporal downscaling, reference-free evaluation, and integration of physical constraints are beyond the scope of this work but represent important directions for future research. In addition, the computational cost of Transformer based architectures increases with global field size, although full-image training experiments demonstrate practical scalability on widely available computing infrastructure. Overall, ViSIR and ViFOR provide a robust and interpretable foundation for frequency-aware climate super-resolution and open promising avenues for extending data-driven downscaling toward spatio-temporal and physics-informed frameworks. Consistent with the statistical nature of learning based super-resolution, the proposed methods do not replace physical or dynamical downscaling but instead provide a complementary, frequency-aware post-processing framework to enhance the spatial fidelity of coarse-resolution ESM outputs.

\section{Conclusion and Future Work}
\label{sec-conclusion}

This work introduces ViSIR and ViFOR as frequency-aware architectures for super-resolution of Earth system model (ESM) outputs, explicitly targeting the fine-scale spatial variability essential for modern geosensing applications. Both models recover small-scale structure that conventional deep learning methods systematically smooth out, with ViFOR consistently achieving the strongest quantitative performance and spectral fidelity, while maintaining favorable scalability under the evaluated training and resolution settings. Future research directions include improving model efficiency through compression and architectural simplification; developing systematic strategies for tuning frequency-related hyperparameters; extending the framework to spatio-temporal super-resolution using temporally ordered inputs; integrating physically consistent loss formulations that preserve key conservation properties; and evaluating scalability on larger, multi-model, and multi-sensor datasets. These advances will be guided by geosensing use cases that require high-resolution, uncertainty-aware products for monitoring hydroclimate extremes, land–water interactions, and other environmental hazards. Within the broader landscape of Transformer based vision models, these results underscore the importance of task-specific, physics- and spectrum-aware architectural choices for climate and geosensing super-resolution, rather than relying on generic image models. By explicitly mitigating spectral bias and enabling scale-aware reconstruction, ViSIR and ViFOR address challenges that general-purpose vision Transformers cannot, providing geoscientists with more reliable and interpretable tools for deriving fine-scale information from coarse-resolution simulations. As statistical post-processing methods, ViSIR and ViFOR are designed to complement, not replace, dynamical downscaling and process-based models, offering a frequency-aware pathway to enhance the spatial fidelity and geosensing utility of climate model outputs and thereby expanding the community's ability to use AI-enhanced products in applications such as hazard mapping, water-resource assessment, and climate-adaptation planning.

\section{Ethics approval and consent to participate}
Ethics approval and consent to participate were not required for this study, as it used only publicly available datasets and did not involve interaction with human or animal subjects.

\section{Competing interest}
The authors declare that they have no competing interests.

\section{Availability of data and material}
The dataset used in this study is publicly available at \cite{E3SM2018} or contact \href{mailto:ezeraatkar@gmail.com}{Ehsan Zeraatkar}.

\section{Funding}
S.A.F. and J. T. would like to acknowledge support by the Department of Energy's Biological and Environmental Research (BER) program (award no. DE-SC0023044)

\bibliographystyle{elsarticle-num-names}
\bibliography{Bib/INR}

\begin{thebibliography}{43}
\expandafter\ifx\csname natexlab\endcsname\relax\def\natexlab#1{#1}\fi
\providecommand{\url}[1]{\texttt{#1}}
\providecommand{\href}[2]{#2}
\providecommand{\path}[1]{#1}
\providecommand{\DOIprefix}{doi:}
\providecommand{\ArXivprefix}{arXiv:}
\providecommand{\URLprefix}{URL: }
\providecommand{\Pubmedprefix}{pmid:}
\providecommand{\doi}[1]{\href{http://dx.doi.org/#1}{\path{#1}}}
\providecommand{\Pubmed}[1]{\href{pmid:#1}{\path{#1}}}
\providecommand{\bibinfo}[2]{#2}
\ifx\xfnm\relax \def\xfnm[#1]{\unskip,\space#1}\fi
\bibitem[{Collins et~al.(2006)Collins, Bitz, Blackmon, Bonan, Bretherton, Carton, Chang, Doney, Hack, Henderson, Kiehl, Large, McKenna, Santer, and Smith}]{collins2006}
\bibinfo{author}{W.~D. Collins}, \bibinfo{author}{C.~M. Bitz}, \bibinfo{author}{M.~L. Blackmon}, \bibinfo{author}{G.~B. Bonan}, \bibinfo{author}{C.~S. Bretherton}, \bibinfo{author}{J.~A. Carton}, \bibinfo{author}{P.~Chang}, \bibinfo{author}{S.~C. Doney}, \bibinfo{author}{J.~J. Hack}, \bibinfo{author}{T.~B. Henderson}, \bibinfo{author}{J.~T. Kiehl}, \bibinfo{author}{W.~G. Large}, \bibinfo{author}{D.~S. McKenna}, \bibinfo{author}{B.~D. Santer}, \bibinfo{author}{R.~D. Smith},
\newblock \bibinfo{title}{The community climate system model version 3 (ccsm3)},
\newblock \bibinfo{journal}{Journal of Climate} \bibinfo{volume}{19} (\bibinfo{year}{2006}) \bibinfo{pages}{2122--2143}. \DOIprefix\doi{10.1175/JCLI3761.1}.
\bibitem[{Heinze et~al.(2019)Heinze, Eyring, Friedlingstein, Jones, Balkanski, Collins, Fichefet, Gao, Hall, Ivanova et~al.}]{heinze2019}
\bibinfo{author}{C.~Heinze}, \bibinfo{author}{V.~Eyring}, \bibinfo{author}{P.~Friedlingstein}, \bibinfo{author}{C.~Jones}, \bibinfo{author}{Y.~Balkanski}, \bibinfo{author}{W.~Collins}, \bibinfo{author}{T.~Fichefet}, \bibinfo{author}{S.~Gao}, \bibinfo{author}{A.~Hall}, \bibinfo{author}{D.~Ivanova}, et~al.,
\newblock \bibinfo{title}{Esd reviews: Climate feedbacks in the earth system and prospects for their evaluation},
\newblock \bibinfo{journal}{Earth Syst. Dynam.} \bibinfo{volume}{10} (\bibinfo{year}{2019}) \bibinfo{pages}{379--452}. \URLprefix \url{https://doi.org/10.5194/esd-10-379-2019}. \DOIprefix\doi{10.5194/esd-10-379-2019}.
\bibitem[{Eyring et~al.(2016)Eyring, Bony, Meehl, Senior, Stevens, Stouffer, and Taylor}]{eyring2016}
\bibinfo{author}{V.~Eyring}, \bibinfo{author}{S.~Bony}, \bibinfo{author}{G.~A. Meehl}, \bibinfo{author}{C.~A. Senior}, \bibinfo{author}{B.~Stevens}, \bibinfo{author}{R.~J. Stouffer}, \bibinfo{author}{K.~E. Taylor},
\newblock \bibinfo{title}{Overview of the coupled model intercomparison project phase 6 (cmip6) experimental design and organization},
\newblock \bibinfo{journal}{Geoscientific Model Development} \bibinfo{volume}{9} (\bibinfo{year}{2016}) \bibinfo{pages}{1937--1958}. \URLprefix \url{https://doi.org/10.5194/gmd-9-1937-2016}. \DOIprefix\doi{10.5194/gmd-9-1937-2016}.
\bibitem[{Rahaman et~al.(2019)Rahaman, Baratin, Arpit, Draxler, Lin, Hamprecht, Bengio, and Courville}]{rahaman2019}
\bibinfo{author}{N.~Rahaman}, \bibinfo{author}{A.~Baratin}, \bibinfo{author}{D.~Arpit}, \bibinfo{author}{F.~Draxler}, \bibinfo{author}{M.~Lin}, \bibinfo{author}{F.~A. Hamprecht}, \bibinfo{author}{Y.~Bengio}, \bibinfo{author}{A.~Courville},
\newblock \bibinfo{title}{On the spectral bias of neural networks},
\newblock in: \bibinfo{booktitle}{International conference on machine learning}, \bibinfo{organization}{PMLR}, \bibinfo{year}{2019}, pp. \bibinfo{pages}{5301--5310}.
\bibitem[{Bashir et~al.(2021)Bashir, Larsen, Ziebell, Fugleholm, and Law}]{Bashir2021}
\bibinfo{author}{A.~Bashir}, \bibinfo{author}{V.~A. Larsen}, \bibinfo{author}{M.~Ziebell}, \bibinfo{author}{K.~Fugleholm}, \bibinfo{author}{I.~Law},
\newblock \bibinfo{title}{Improved detection of postoperative residual meningioma with [68ga]ga-dota-toc pet imaging using a high-resolution research tomograph pet scanner},
\newblock \bibinfo{journal}{Clinical Cancer Research} \bibinfo{volume}{27} (\bibinfo{year}{2021}) \bibinfo{pages}{2216--2225}. \URLprefix \url{https://doi.org/10.1158/1078-0432.CCR-20-3362}. \DOIprefix\doi{10.1158/1078-0432.CCR-20-3362}.
\bibitem[{Kuzmanić et~al.(2007)Kuzmanić, Vujović, and Vujović}]{kuzmanic2007}
\bibinfo{author}{I.~Kuzmanić}, \bibinfo{author}{I.~Vujović}, \bibinfo{author}{M.~Vujović},
\newblock \bibinfo{title}{Application of computer vision in security and emergency actions},
\newblock in: \bibinfo{booktitle}{Proceedings of the 14th Annual Conference of the International Emergency Management Society}, \bibinfo{publisher}{TIEMS}, \bibinfo{address}{Split}, \bibinfo{year}{2007}, pp. \bibinfo{pages}{336--345--x}.
\bibitem[{Grosche et~al.(2021)Grosche, Brand, and Kaup}]{Grosche2021}
\bibinfo{author}{S.~Grosche}, \bibinfo{author}{F.~Brand}, \bibinfo{author}{A.~Kaup},
\newblock \bibinfo{title}{A novel end-to-end network for reconstruction of non-regularly sampled image data using locally fully connected layers},
\newblock in: \bibinfo{booktitle}{2021 IEEE 23rd International Workshop on Multimedia Signal Processing (MMSP)}, \bibinfo{year}{2021}, pp. \bibinfo{pages}{1--6}. \DOIprefix\doi{10.1109/MMSP53017.2021.9733541}.
\bibitem[{Maral(2022)}]{Maral2022}
\bibinfo{author}{B.~C. Maral},
\newblock \bibinfo{title}{Single image super-resolution methods: A survey},
\newblock \bibinfo{journal}{https://doi.org/10.48550/arXiv.2202.11763}  (\bibinfo{year}{2022}).
\bibitem[{Ledig et~al.(2017)Ledig, Theis, Husz{\'a}r, Caballero, Cunningham, Acosta, Aitken, Tejani, Totz, Wang, and Shi}]{ledig2017}
\bibinfo{author}{C.~Ledig}, \bibinfo{author}{L.~Theis}, \bibinfo{author}{F.~Husz{\'a}r}, \bibinfo{author}{J.~Caballero}, \bibinfo{author}{A.~Cunningham}, \bibinfo{author}{A.~Acosta}, \bibinfo{author}{A.~Aitken}, \bibinfo{author}{A.~Tejani}, \bibinfo{author}{J.~Totz}, \bibinfo{author}{Z.~Wang}, \bibinfo{author}{W.~Shi},
\newblock \bibinfo{title}{Photo-realistic single image super-resolution using a generative adversarial network},
\newblock in: \bibinfo{booktitle}{Proceedings of the IEEE Conference on Computer Vision and Pattern Recognition (CVPR)}, \bibinfo{year}{2017}, pp. \bibinfo{pages}{4681--4690}. \URLprefix \url{https://arxiv.org/abs/1609.04802}.
\bibitem[{Lim et~al.(2017)Lim, Son, Kim, Nah, and Lee}]{lim2017}
\bibinfo{author}{B.~Lim}, \bibinfo{author}{S.~Son}, \bibinfo{author}{H.~Kim}, \bibinfo{author}{S.~Nah}, \bibinfo{author}{K.~M. Lee},
\newblock \bibinfo{title}{Enhanced deep residual networks for single image super-resolution},
\newblock \bibinfo{journal}{2017 IEEE Conference on Computer Vision and Pattern Recognition Workshops (CVPRW)}  (\bibinfo{year}{2017}) \bibinfo{pages}{1132--1140}. \URLprefix \url{https://api.semanticscholar.org/CorpusID:6540453}.
\bibitem[{Vandal et~al.(2017)Vandal, Kodra, and Ganguly}]{vandal2017}
\bibinfo{author}{T.~Vandal}, \bibinfo{author}{E.~Kodra}, \bibinfo{author}{A.~R. Ganguly},
\newblock \bibinfo{title}{Deepsd: Generating high resolution climate change projections through single image super-resolution},
\newblock \bibinfo{journal}{arXiv preprint arXiv:1703.03126}  (\bibinfo{year}{2017}). \URLprefix \url{https://arxiv.org/abs/1703.03126}, \bibinfo{note}{accessed: 2024-09-02}.
\bibitem[{Kim et~al.(2016)Kim, Lee, and Lee}]{Kim2016VDSR}
\bibinfo{author}{J.~Kim}, \bibinfo{author}{J.~K. Lee}, \bibinfo{author}{K.~M. Lee},
\newblock \bibinfo{title}{Accurate image super-resolution using very deep convolutional networks},
\newblock in: \bibinfo{booktitle}{2016 IEEE Conference on Computer Vision and Pattern Recognition (CVPR)}, \bibinfo{year}{2016}, pp. \bibinfo{pages}{1646--1654}. \DOIprefix\doi{10.1109/CVPR.2016.182}.
\bibitem[{Dosovitskiy et~al.(2020)Dosovitskiy, Beyer, Kolesnikov, Weissenborn, Zhai, Unterthiner, Dehghani, Minderer, Heigold, Gelly, Uszkoreit, and Houlsby}]{Dosovitskiy2020}
\bibinfo{author}{A.~Dosovitskiy}, \bibinfo{author}{L.~Beyer}, \bibinfo{author}{A.~Kolesnikov}, \bibinfo{author}{D.~Weissenborn}, \bibinfo{author}{X.~Zhai}, \bibinfo{author}{T.~Unterthiner}, \bibinfo{author}{M.~Dehghani}, \bibinfo{author}{M.~Minderer}, \bibinfo{author}{G.~Heigold}, \bibinfo{author}{S.~Gelly}, \bibinfo{author}{J.~Uszkoreit}, \bibinfo{author}{N.~Houlsby},
\newblock \bibinfo{title}{An image is worth 16x16 words: Transformers for image recognition at scale},
\newblock \bibinfo{journal}{ArXiv} \bibinfo{volume}{abs/2010.11929} (\bibinfo{year}{2020}). \URLprefix \url{https://api.semanticscholar.org/CorpusID:225039882}.
\bibitem[{Sohl-Dickstein et~al.(2015)Sohl-Dickstein, Weiss, Maheswaranathan, and Ganguli}]{Dickstein2015}
\bibinfo{author}{J.~Sohl-Dickstein}, \bibinfo{author}{E.~A. Weiss}, \bibinfo{author}{N.~Maheswaranathan}, \bibinfo{author}{S.~Ganguli},
\newblock \bibinfo{title}{Deep unsupervised learning using nonequilibrium thermodynamics},
\newblock in: \bibinfo{booktitle}{Proceedings of the 32nd International Conference on International Conference on Machine Learning - Volume 37}, ICML'15, \bibinfo{publisher}{JMLR.org}, \bibinfo{year}{2015}, p. \bibinfo{pages}{2256–2265}.
\bibitem[{Bai et~al.(2022)Bai, Yuan, Xia, Yan, Li, and Liu}]{bai2022improving}
\bibinfo{author}{J.~Bai}, \bibinfo{author}{L.~Yuan}, \bibinfo{author}{S.-T. Xia}, \bibinfo{author}{S.~Yan}, \bibinfo{author}{Z.~Li}, \bibinfo{author}{W.~Liu},
\newblock \bibinfo{title}{Improving vision transformers by revisiting high-frequency components},
\newblock in: \bibinfo{booktitle}{European Conference on Computer Vision}, \bibinfo{organization}{Springer}, \bibinfo{year}{2022}, pp. \bibinfo{pages}{1--18}.
\bibitem[{Yang et~al.(2010)Yang, Wright, Huang, and Ma}]{Yang2010}
\bibinfo{author}{J.~Yang}, \bibinfo{author}{J.~Wright}, \bibinfo{author}{T.~S. Huang}, \bibinfo{author}{Y.~Ma},
\newblock \bibinfo{title}{Image super-resolution via sparse representation},
\newblock \bibinfo{journal}{IEEE Transactions on Image Processing} \bibinfo{volume}{19} (\bibinfo{year}{2010}) \bibinfo{pages}{2861--2873}. \DOIprefix\doi{10.1109/TIP.2010.2050625}.
\bibitem[{Dong et~al.(2016)Dong, Loy, He, and Tang}]{Dong2016}
\bibinfo{author}{C.~Dong}, \bibinfo{author}{C.~C. Loy}, \bibinfo{author}{K.~He}, \bibinfo{author}{X.~Tang},
\newblock \bibinfo{title}{Image super-resolution using deep convolutional networks},
\newblock \bibinfo{journal}{IEEE Transactions on Pattern Analysis and Machine Intelligence} \bibinfo{volume}{38} (\bibinfo{year}{2016}) \bibinfo{pages}{295--307}. \DOIprefix\doi{10.1109/TPAMI.2015.2439281}.
\bibitem[{Kim et~al.(2016)Kim, Lee, and Lee}]{Kim2016EDSR}
\bibinfo{author}{J.~Kim}, \bibinfo{author}{J.~K. Lee}, \bibinfo{author}{K.~M. Lee},
\newblock \bibinfo{title}{Enhanced deep residual networks for single image super-resolution},
\newblock \bibinfo{journal}{Proceedings of the IEEE Conference on Computer Vision and Pattern Recognition (CVPR)}  (\bibinfo{year}{2016}) \bibinfo{pages}{136--144}.
\bibitem[{Zhang et~al.(2018)Zhang, Tian, Kong, Zhong, and Fu}]{Zhang2018}
\bibinfo{author}{Y.~Zhang}, \bibinfo{author}{Y.~Tian}, \bibinfo{author}{Y.~Kong}, \bibinfo{author}{B.~Zhong}, \bibinfo{author}{Y.~Fu},
\newblock \bibinfo{title}{Residual dense network for image super-resolution},
\newblock in: \bibinfo{booktitle}{2018 IEEE/CVF Conference on Computer Vision and Pattern Recognition}, \bibinfo{year}{2018}, pp. \bibinfo{pages}{2472--2481}. \DOIprefix\doi{10.1109/CVPR.2018.00262}.
\bibitem[{Zhang et~al.(2019)Zhang, Zhang, Wu, and Zhang}]{Zhang2019}
\bibinfo{author}{X.~Zhang}, \bibinfo{author}{Z.~Zhang}, \bibinfo{author}{S.~Wu}, \bibinfo{author}{Z.~Zhang},
\newblock \bibinfo{title}{Residual networks behave like ensembles of relatively shallow networks},
\newblock \bibinfo{journal}{IEEE Transactions on Pattern Analysis and Machine Intelligence} \bibinfo{volume}{41} (\bibinfo{year}{2019}) \bibinfo{pages}{1311--1325}.
\bibitem[{Zhu et~al.(2025{\natexlab{a}})Zhu, Deng, Song, Li, and Wang}]{Mamba2025}
\bibinfo{author}{C.~Zhu}, \bibinfo{author}{S.~Deng}, \bibinfo{author}{X.~Song}, \bibinfo{author}{Y.~Li}, \bibinfo{author}{Q.~Wang},
\newblock \bibinfo{title}{Mamba collaborative implicit neural representation for hyperspectral and multispectral remote sensing image fusion},
\newblock \bibinfo{journal}{IEEE Transactions on Geoscience and Remote Sensing} \bibinfo{volume}{63} (\bibinfo{year}{2025}{\natexlab{a}}) \bibinfo{pages}{1--15}. \DOIprefix\doi{10.1109/TGRS.2025.3537638}.
\bibitem[{Zhu et~al.(2025{\natexlab{b}})Zhu, Song, Li, Deng, and Zhang}]{ZHU2025}
\bibinfo{author}{C.~Zhu}, \bibinfo{author}{X.~Song}, \bibinfo{author}{Y.~Li}, \bibinfo{author}{S.~Deng}, \bibinfo{author}{T.~Zhang},
\newblock \bibinfo{title}{A spatial-frequency dual-domain implicit guidance method for hyperspectral and multispectral remote sensing image fusion based on kolmogorov–arnold network},
\newblock \bibinfo{journal}{Information Fusion} \bibinfo{volume}{123} (\bibinfo{year}{2025}{\natexlab{b}}) \bibinfo{pages}{103261}. \DOIprefix\doi{https://doi.org/10.1016/j.inffus.2025.103261}.
\bibitem[{Sitzmann et~al.(2020)Sitzmann, Martel, Bergman, Lindell, and Wetzstein}]{SIREN}
\bibinfo{author}{V.~Sitzmann}, \bibinfo{author}{J.~N.~P. Martel}, \bibinfo{author}{A.~W. Bergman}, \bibinfo{author}{D.~B. Lindell}, \bibinfo{author}{G.~Wetzstein},
\newblock \bibinfo{title}{Implicit neural representations with periodic activation functions},
\newblock in: \bibinfo{booktitle}{Proceedings of the 34th International Conference on Neural Information Processing Systems}, volume~\bibinfo{volume}{33} of \textit{\bibinfo{series}{NIPS '20}}, \bibinfo{publisher}{Curran Associates Inc.}, \bibinfo{address}{Red Hook, NY, USA}, \bibinfo{year}{2020}, pp. \bibinfo{pages}{7462--7473}.
\bibitem[{Grattarola and Vandergheynst(2022)}]{grattarola2022ginr}
\bibinfo{author}{D.~Grattarola}, \bibinfo{author}{P.~Vandergheynst},
\newblock \bibinfo{title}{Generalised implicit neural representations},
\newblock in: \bibinfo{booktitle}{Proceedings of the 36th International Conference on Neural Information Processing Systems}, volume~\bibinfo{volume}{35} of \textit{\bibinfo{series}{NIPS '22}}, \bibinfo{publisher}{Curran Associates Inc.}, \bibinfo{address}{Red Hook, NY, USA}, \bibinfo{year}{2022}, pp. \bibinfo{pages}{30446--30458}.
\bibitem[{Chen et~al.(2024)Chen, Wu, Liu, and Zhu}]{chen2024hoin}
\bibinfo{author}{Y.~Chen}, \bibinfo{author}{R.~Wu}, \bibinfo{author}{Y.~Liu}, \bibinfo{author}{C.~Zhu}, \bibinfo{title}{Hoin: High-order implicit neural representations}, \bibinfo{year}{2024}. \URLprefix \url{https://arxiv.org/abs/2404.14674}. \href{http://arxiv.org/abs/2404.14674}{{\tt arXiv:2404.14674}}.
\bibitem[{Shidqi et~al.(2023)Shidqi, Jeong, Park, Zeller, Nellikkattil, and Singh}]{shidqi2023}
\bibinfo{author}{N.~Shidqi}, \bibinfo{author}{C.~Jeong}, \bibinfo{author}{S.~Park}, \bibinfo{author}{E.~Zeller}, \bibinfo{author}{A.~B. Nellikkattil}, \bibinfo{author}{K.~Singh}, \bibinfo{title}{Generating high-resolution regional precipitation using conditional diffusion model}, \bibinfo{year}{2023}. \URLprefix \url{https://arxiv.org/abs/2312.07112}. \href{http://arxiv.org/abs/2312.07112}{{\tt arXiv:2312.07112}}.
\bibitem[{Ding et~al.(2024)Ding, Zhi, Lyu, Ji, and Guo}]{Ding2024}
\bibinfo{author}{S.~Ding}, \bibinfo{author}{X.~Zhi}, \bibinfo{author}{Y.~Lyu}, \bibinfo{author}{Y.~Ji}, \bibinfo{author}{W.~Guo},
\newblock \bibinfo{title}{Deep learning for daily 2‐m temperature downscaling},
\newblock \bibinfo{journal}{Earth and Space Science} \bibinfo{volume}{11} (\bibinfo{year}{2024}). \DOIprefix\doi{10.1029/2023EA003227}.
\bibitem[{Zhang et~al.(2025)Zhang, Liu, Chen, Li, and Wang}]{neurOpDiff2025}
\bibinfo{author}{Z.~Zhang}, \bibinfo{author}{W.~Liu}, \bibinfo{author}{H.~Chen}, \bibinfo{author}{M.~Li}, \bibinfo{author}{Y.~Wang},
\newblock \bibinfo{title}{Neurop-diff: Continuous remote sensing image super-resolution via neural operator diffusion},
\newblock \bibinfo{journal}{arXiv preprint arXiv:2501.09054}  (\bibinfo{year}{2025}). \URLprefix \url{https://arxiv.org/abs/2501.09054}, \bibinfo{note}{code available at \url{https://github.com/zerono000/NeurOp-Diff}}.
\bibitem[{Kumar et~al.(2025)Kumar, Li, Zhao, and Wang}]{pcSRGAN2025}
\bibinfo{author}{A.~Kumar}, \bibinfo{author}{J.~Li}, \bibinfo{author}{R.~Zhao}, \bibinfo{author}{X.~Wang},
\newblock \bibinfo{title}{Pc-srgan: Physically consistent super-resolution generative adversarial network for general transient simulations},
\newblock \bibinfo{journal}{arXiv preprint arXiv:2505.06502}  (\bibinfo{year}{2025}). \URLprefix \url{https://arxiv.org/abs/2505.06502}.
\bibitem[{Zhong et~al.(2023)Zhong, Du, Chen, Wang, and Li}]{Zhong2023}
\bibinfo{author}{X.~Zhong}, \bibinfo{author}{F.~Du}, \bibinfo{author}{L.~Chen}, \bibinfo{author}{Z.~Wang}, \bibinfo{author}{H.~Li},
\newblock \bibinfo{title}{Investigating transformer‐based models for spatial downscaling and correcting biases of near‐surface temperature and wind‐speed forecasts},
\newblock \bibinfo{journal}{Quarterly Journal of the Royal Meteorological Society} \bibinfo{volume}{150} (\bibinfo{year}{2023}) \bibinfo{pages}{275--289}. \DOIprefix\doi{10.1002/qj.4596}.
\bibitem[{Karwowska and Wierzbicki(2024)}]{Karwowska2024}
\bibinfo{author}{K.~Karwowska}, \bibinfo{author}{D.~Wierzbicki},
\newblock \bibinfo{title}{Modified esrgan with uformer for video satellite imagery super-resolution},
\newblock \bibinfo{journal}{Remote Sensing} \bibinfo{volume}{16} (\bibinfo{year}{2024}) \bibinfo{pages}{1926}. \DOIprefix\doi{10.3390/rs16111926}.
\bibitem[{Liu et~al.(2025)Liu, Wang, Hu, and Chen}]{ttrd32025}
\bibinfo{author}{S.~Liu}, \bibinfo{author}{K.~Wang}, \bibinfo{author}{Z.~Hu}, \bibinfo{author}{L.~Chen},
\newblock \bibinfo{title}{Ttrd3: Texture transfer residual denoising dual diffusion model for remote sensing image super-resolution},
\newblock \bibinfo{journal}{arXiv preprint arXiv:2504.13026}  (\bibinfo{year}{2025}). \URLprefix \url{https://arxiv.org/abs/2504.13026}.
\bibitem[{Hewitson and Crane(1996)}]{hewitson1996climate}
\bibinfo{author}{B.~C. Hewitson}, \bibinfo{author}{R.~G. Crane},
\newblock \bibinfo{title}{Climate downscaling: Techniques and application},
\newblock \bibinfo{journal}{Climate Research} \bibinfo{volume}{7} (\bibinfo{year}{1996}) \bibinfo{pages}{85--95}. \DOIprefix\doi{10.3354/cr007085}.
\bibitem[{Hewitson et~al.(2014)Hewitson, Daron, Crane et~al.}]{hewitson2014interrogating}
\bibinfo{author}{B.~C. Hewitson}, \bibinfo{author}{J.~Daron}, \bibinfo{author}{R.~G. Crane}, et~al.,
\newblock \bibinfo{title}{Interrogating empirical--statistical downscaling},
\newblock \bibinfo{journal}{Climatic Change} \bibinfo{volume}{122} (\bibinfo{year}{2014}) \bibinfo{pages}{539--554}. \DOIprefix\doi{10.1007/s10584-013-1021-z}.
\bibitem[{Lantuejoul(2002)}]{lantuejoul2002geostatistical}
\bibinfo{author}{C.~Lantuejoul}, \bibinfo{title}{Geostatistical Simulation}, \bibinfo{publisher}{Springer}, \bibinfo{year}{2002}. \DOIprefix\doi{10.1007/978-3-662-04808-5}.
\bibitem[{Carion et~al.(2020)Carion, Massa, Synnaeve, Usunier, Kirillov, and Zagoruyko}]{carion2020end}
\bibinfo{author}{N.~Carion}, \bibinfo{author}{F.~Massa}, \bibinfo{author}{G.~Synnaeve}, \bibinfo{author}{N.~Usunier}, \bibinfo{author}{A.~Kirillov}, \bibinfo{author}{S.~Zagoruyko},
\newblock \bibinfo{title}{End-to-end object detection with transformers},
\newblock in: \bibinfo{booktitle}{European Conference on Computer Vision (ECCV)}, \bibinfo{year}{2020}, pp. \bibinfo{pages}{213--229}.
\bibitem[{Liu et~al.(2021)Liu, Lin, Cao, Hu, Wei, Zhang, Lin, and Guo}]{liu2021swin}
\bibinfo{author}{Z.~Liu}, \bibinfo{author}{Y.~Lin}, \bibinfo{author}{Y.~Cao}, \bibinfo{author}{H.~Hu}, \bibinfo{author}{Y.~Wei}, \bibinfo{author}{Z.~Zhang}, \bibinfo{author}{S.~Lin}, \bibinfo{author}{B.~Guo},
\newblock \bibinfo{title}{Swin transformer: Hierarchical vision transformer using shifted windows},
\newblock in: \bibinfo{booktitle}{Proceedings of the IEEE/CVF international conference on computer vision}, \bibinfo{year}{2021}, pp. \bibinfo{pages}{10012--10022}.
\bibitem[{He et~al.(2021)He, Chen, Xie, Li, Dollár, and Girshick}]{he2022masked}
\bibinfo{author}{K.~He}, \bibinfo{author}{X.~Chen}, \bibinfo{author}{S.~Xie}, \bibinfo{author}{Y.~Li}, \bibinfo{author}{P.~Dollár}, \bibinfo{author}{R.~Girshick}, \bibinfo{title}{Masked autoencoders are scalable vision learners}, \bibinfo{year}{2021}. \URLprefix \url{https://arxiv.org/abs/2111.06377}. \href{http://arxiv.org/abs/2111.06377}{{\tt arXiv:2111.06377}}.
\bibitem[{Passarella et~al.(2022)Passarella, Mahajan, Pal, and Norman}]{Passarella2022}
\bibinfo{author}{L.~S. Passarella}, \bibinfo{author}{S.~Mahajan}, \bibinfo{author}{A.~Pal}, \bibinfo{author}{M.~R. Norman},
\newblock \bibinfo{title}{Reconstructing high-resolution esm data through a novel fast super-resolution convolutional neural network (fsrcnn)},
\newblock \bibinfo{journal}{Earth and Space Science} \bibinfo{volume}{49} (\bibinfo{year}{2022}). \DOIprefix\doi{10.1029/2021GL097571}.
\bibitem[{Wang et~al.(2004)Wang, Bovik, Sheikh, and Simoncelli}]{SSIM}
\bibinfo{author}{Z.~Wang}, \bibinfo{author}{A.~Bovik}, \bibinfo{author}{H.~Sheikh}, \bibinfo{author}{E.~Simoncelli},
\newblock \bibinfo{title}{Image quality assessment: from error visibility to structural similarity},
\newblock \bibinfo{journal}{IEEE Transactions on Image Processing} \bibinfo{volume}{13} (\bibinfo{year}{2004}) \bibinfo{pages}{600--612}. \DOIprefix\doi{10.1109/TIP.2003.819861}.
\bibitem[{Vandal et~al.(2017)Vandal, Kodra, Ganguly, Michaelis, Nemani, and Ganguly}]{vandal2017deepsd}
\bibinfo{author}{T.~Vandal}, \bibinfo{author}{E.~Kodra}, \bibinfo{author}{S.~Ganguly}, \bibinfo{author}{A.~Michaelis}, \bibinfo{author}{R.~Nemani}, \bibinfo{author}{A.~R. Ganguly},
\newblock \bibinfo{title}{Deepsd: Generating high resolution climate change projections through single image super-resolution},
\newblock \bibinfo{journal}{Proceedings of the 23rd ACM SIGKDD}  (\bibinfo{year}{2017}).
\bibitem[{Stengel et~al.(2020)Stengel, Glaws, Hettinger, and King}]{stengel2020adversarial}
\bibinfo{author}{K.~Stengel}, \bibinfo{author}{A.~Glaws}, \bibinfo{author}{D.~Hettinger}, \bibinfo{author}{R.~King},
\newblock \bibinfo{title}{Adversarial super-resolution of climatological wind and solar data},
\newblock \bibinfo{journal}{Proceedings of the National Academy of Sciences}  (\bibinfo{year}{2020}).
\bibitem[{{E3SM Project}(2024)}]{E3SM2018}
\bibinfo{author}{{E3SM Project}}, \bibinfo{title}{{Energy Exascale Earth System Model (E3SM)}}, \bibinfo{howpublished}{[Computer Software]}, \bibinfo{year}{2024}. \DOIprefix\doi{10.11578/E3SM/dc.20240301.3}.

\end{thebibliography}

\end{document}